\documentclass[10pt,twocolumn,letterpaper]{article}

\usepackage[pagenumbers]{cvpr}

\usepackage{cuted} %

\usepackage{currfile} %

\usepackage{caption} %

\usepackage{xspace}
\usepackage{enumitem}
\usepackage{booktabs}
\usepackage{makecell}
\usepackage{multirow}
\usepackage{pifont}
\usepackage{graphicx}

\usepackage{array}
\usepackage{colortbl}
\usepackage{amssymb}
\usepackage{tikz}

\definecolor{secblue}{HTML}{EEF4FB}\definecolor{peach}{HTML}{FDEBDD}\definecolor{ylw}{HTML}{FDF6D8}
\definecolor{dkg}{HTML}{9E9E9E}\definecolor{ltg}{HTML}{E6E6E6}
\definecolor{g1}{HTML}{4CD964}\definecolor{g2}{HTML}{A8EFA0}\definecolor{g3}{HTML}{D9F7D3}
\definecolor{cBase}{HTML}{5B6B7C}\definecolor{cSpa}{HTML}{3D76B0}\definecolor{cTem}{HTML}{C9763A}\definecolor{cDyn}{HTML}{3F936B}\definecolor{cApp}{HTML}{A88A2A}

\newcommand{\lgO}{\raisebox{-0.12em}{\includegraphics[height=0.85em]{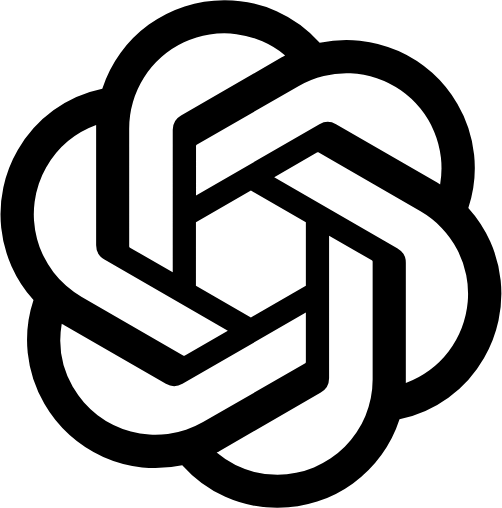}}\hspace{0.28em}}
\newcommand{\lgC}{\raisebox{-0.12em}{\includegraphics[height=0.85em]{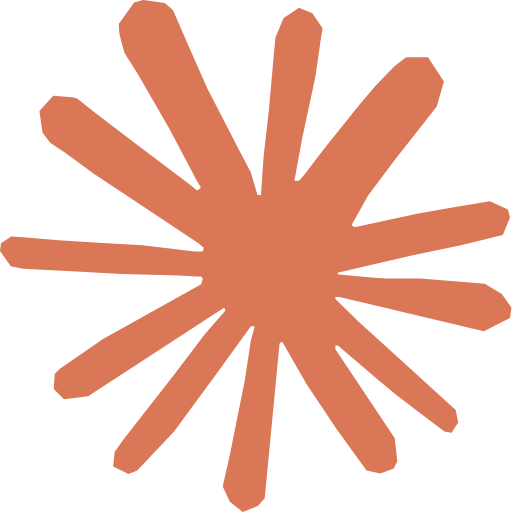}}\hspace{0.28em}}
\newcommand{\lgMM}{\raisebox{-0.12em}{\includegraphics[height=0.85em]{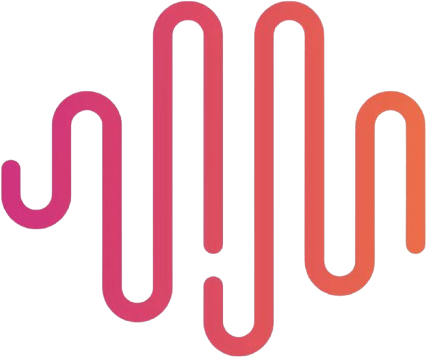}}\hspace{0.28em}}
\newcommand{\lgSD}{\raisebox{-0.12em}{\includegraphics[height=0.85em]{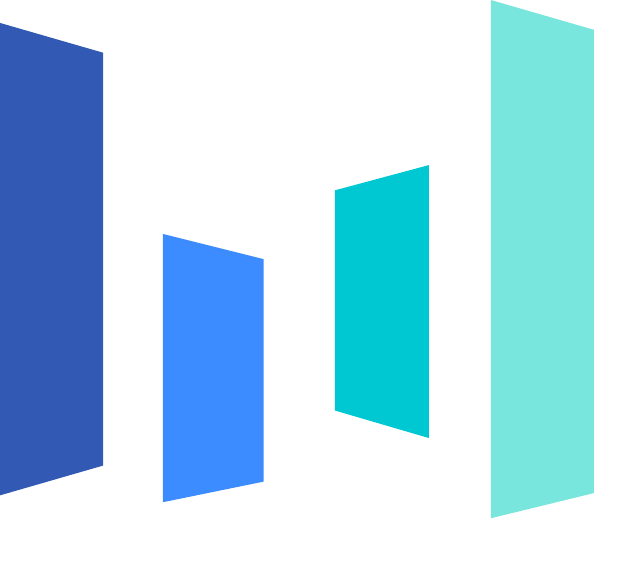}}\hspace{0.28em}}
\newcommand{\lgKL}{\raisebox{-0.12em}{\includegraphics[height=0.85em]{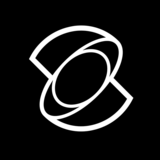}}\hspace{0.28em}}
\definecolor{oHelp}{HTML}{4B5BAB}\definecolor{oI2V}{HTML}{B9C6EE}\definecolor{oLost}{HTML}{C44E62}\definecolor{oMis}{HTML}{F4C3CC}\definecolor{oUns}{HTML}{DCDCDC}
\newcommand{\oc}[2]{\textcolor{#1}{\rule[-0.05em]{0.62em}{0.62em}}\,\emph{#2}}
\newcommand{\catbar}[2]{\begin{tabular}[b]{@{}c@{}}\textcolor{#1}{#2}\\[-3.2pt]\textcolor{#1}{\rule{2.1em}{1.3pt}}\end{tabular}}
\newsavebox{\tabbox}

\usepackage[most]{tcolorbox}
\definecolor{findborder}{HTML}{1F5C5C}
\newtcolorbox{findingbox}{enhanced, colback=gray!8, colframe=findborder, boxrule=0.6pt, arc=2.5mm,
  left=5pt, right=5pt, top=3pt, bottom=3pt, before skip=6pt, after skip=6pt,
  drop shadow={opacity=0.25}, fontupper=\itshape\small}

\definecolor{cvprblue}{rgb}{0.21,0.49,0.74}
\usepackage[pagebackref,breaklinks,colorlinks,allcolors=cvprblue]{hyperref}

\tcbset{promptstyle/.style={enhanced, colback=gray!8, colframe=black, coltitle=white,
  fonttitle=\bfseries\small, fontupper=\small, fontlower=\small, arc=2mm, boxrule=0.8pt,
  left=5pt, right=5pt, top=4pt, bottom=4pt, toptitle=2pt, bottomtitle=2pt,
  before skip=6pt, after skip=6pt, before upper={\setlength{\parskip}{2pt}}, before lower={\setlength{\parskip}{2pt}}, lowerbox=visible, segmentation style={black!40, dashed}}}
\newtcolorbox{promptbox}[2][]{promptstyle, float=tbp, title={#2}, #1} %
\newtcolorbox{promptboxwide}[2][]{promptstyle, title={#2}, #1}
\usepackage{placeins}

\definecolor{boxCon}{HTML}{4A6A8F}   %
\definecolor{boxTtoC}{HTML}{2E7D6E}  %
\definecolor{boxCtoV}{HTML}{6B4E96}  %
\definecolor{boxEval}{HTML}{8C5E24}  %
\tcbset{stageCon/.style={colframe=boxCon, colback=boxCon!7, before upper app={\def\boxcol{boxCon}}},
  stageTtoC/.style={colframe=boxTtoC, colback=boxTtoC!7, before upper app={\def\boxcol{boxTtoC}}},
  stageCtoV/.style={colframe=boxCtoV, colback=boxCtoV!7, before upper app={\def\boxcol{boxCtoV}}, before lower app={\def\boxcol{boxCtoV}}, segmentation style={boxCtoV!50, dashed}},
  stageEval/.style={colframe=boxEval, colback=boxEval!7, before upper app={\def\boxcol{boxEval}}}}
\def\boxcol{black}
\newcommand{\plabel}[1]{\textbf{\textcolor{\boxcol}{[#1]}}}

\title{ClayBench: Evaluating Text-to-Clay-to-Video Generation}

\author{
Hohin Kwan$^{1}$\quad
Xianghao Kong$^{1}$\quad
Songlin Yang$^{1}$\quad
Xuyi Yang$^{1}$\\
Zeyu Zhang$^{1}$\quad
Lvmin Zhang$^{2}$\quad
Anyi Rao$^{1,\dagger}$\\[0.5em]
{\normalsize $^{1}$HKUST MMLab\qquad $^{2}$Stanford University}\\[0.2em]
{\normalsize $^{\dagger}$Corresponding author}\\[0.4em]
{\normalsize\url{https://mrakas.github.io/claybench}}
}

\begin{document}
\maketitle
\begin{strip}
\centering
\includegraphics[width=\linewidth,trim=0 135 0 0,clip]{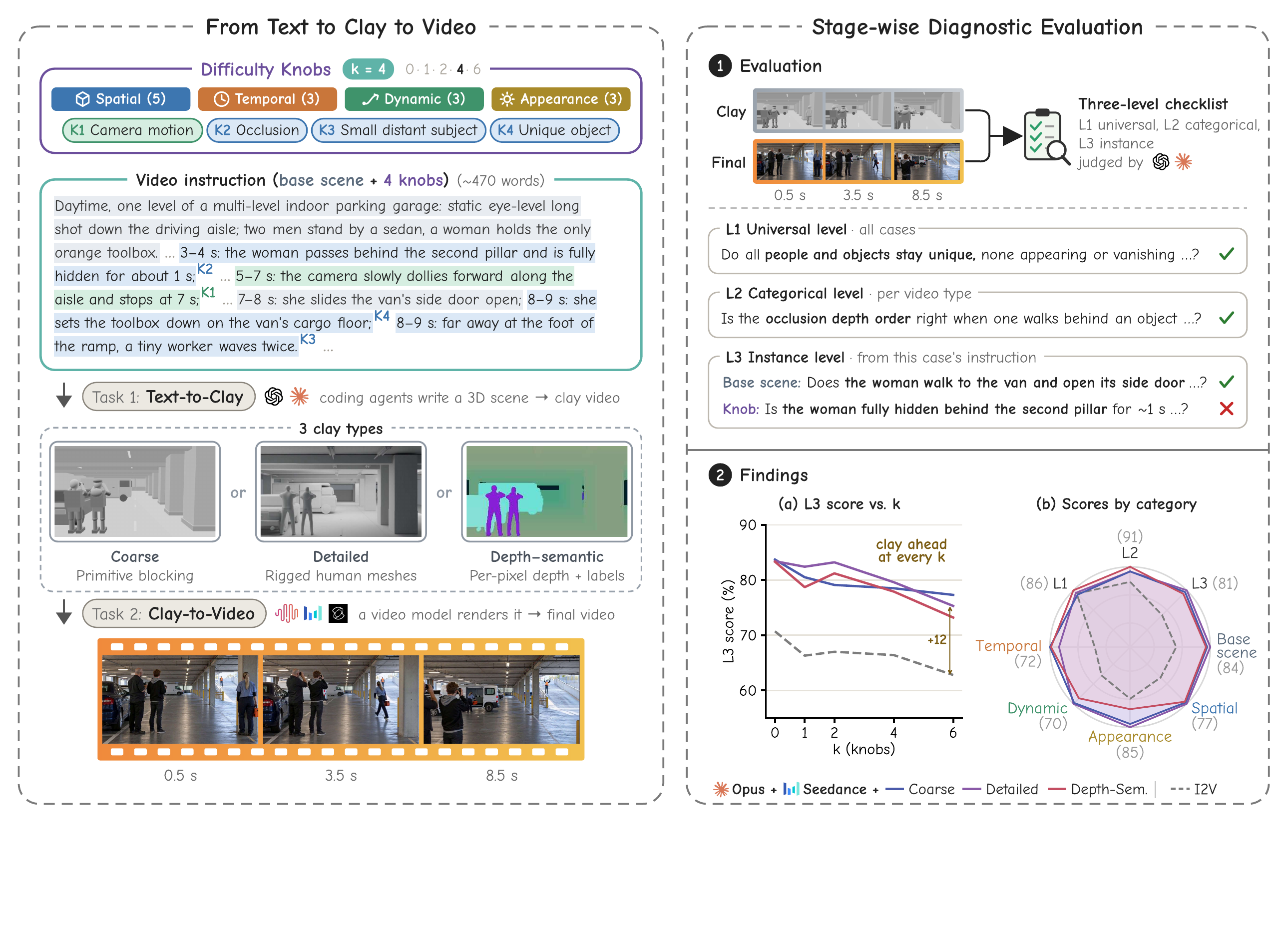}
\captionsetup{hypcap=false}
\captionof{figure}{\textbf{Overview of ClayBench.} We study the two-stage Text-to-Clay-to-Video pipeline: a coding agent builds a 3D clay video from the instruction (Task~1: Text-to-Clay), and a video generation model renders it into pixels (Task~2: Clay-to-Video). Both stages are scored by a unified three-level checklist across $k$ difficulty knobs. Findings: (a) L3 score vs.\ $k$: clay leads image-to-video by 10--16 points at every $k$. (b) Scores by category: clay improves over image-to-video on all four categories, \textcolor{cSpa}{Spatial}, \textcolor{cTem}{Temporal}, \textcolor{cDyn}{Dynamic}, and \textcolor{cApp}{Appearance}. Results use Claude Opus~5.5 clays and Seedance~2.0.}

\captionsetup{hypcap=true}
\label{fig:teaser}%
\end{strip}

\begin{abstract}
Video generation models synthesize photorealistic shots, yet their 4D geometry frequently deviates from complex instructions governing subject staging, occlusion order, camera motion, and event timing. A two-stage code-then-paint pipeline addresses this gap: a coding agent first constructs an untextured 3D layout (Text-to-Clay, T2C), which a video generation model then renders into pixels (Clay-to-Video, C2V). When this pipeline fails, however, existing benchmarks cannot trace whether the error arises in T2C or C2V. We introduce ClayBench, an end-to-end diagnostic benchmark that disentangles both stages by scoring the clay videos and the final videos against a unified three-level checklist (Universal, Categorical, and Instance). Derived from real shots across four video types, ClayBench controls complexity via $k$ difficulty knobs and compares Coarse, Detailed, and Depth-Semantic clay representations. Benchmarking three coding agents and three video generation models yields three key insights. First, generative rendering is a major bottleneck: video generation models lose 29 to 45\% of the checks satisfied in the clay, shrinking the T2C quality gaps across coding agents after rendering, with Dynamic checks lost mainly in C2V and Temporal checks failing mainly in T2C. Second, clay conditioning outperforms Image-to-Video (I2V) at every difficulty level $k$. Third, higher clay scores do not guarantee better final videos: video generation models repair many clay errors, Depth-Semantic clay reaches the highest Overall despite the lowest T2C score, and the best representation depends on the video generation model and video type.
\end{abstract}

\section{Introduction}
\label{sec:intro}

Frontier video generation models such as MiniMax-H3~\cite{minimax2026h3}, Seedance 2.0~\cite{seedance2026}, and Kling 4.0~\cite{kuaishou2026kling4} now synthesize photorealistic shots, yet they remain difficult to steer when an instruction specifies complex 4D geometry: where multiple subjects move, how they occlude one another, when sequential or simultaneous events unfold, and how the camera moves. A shot in which the camera dollies past a waiter who walks behind a pillar before a guest raises a glass often comes out with a static camera, a missing occlusion, or reversed events~\cite{huang2024vbench,zheng2025vbench2,sun2025t2vcompbench}. Filmmakers solve this choreography problem through \emph{previsualization} (\emph{previs}): blocking out staging, timing, and camera motion with untextured 3D geometry (\emph{clay}) before final rendering. Coding agents follow explicit spatial and temporal specifications more faithfully than diffusion models, which motivates a \textbf{Text-to-Clay-to-Video} (\emph{code-then-paint}) pipeline: a coding agent translates the instruction into an executable 3D clay scene (Text-to-Clay, \texttt{T2C}), and a video generation model renders it into pixels (Clay-to-Video, \texttt{C2V}) (\cref{fig:teaser}).

Despite growing interest in code-driven 3D/4D scene generation~\cite{wang2026cinerecon,shen20264dcodebench,fang2026blenderworld,diao2026vibevideobench} and proxy-conditioned video rendering~\cite{chen2026codeworld,liu2026anygamerenderer,huang2026prowbench,qiu2026renderbench,xu2026proxy2world}, a diagnostic question remains open: when does clay help video generation, and where does the Text-to-Clay-to-Video pipeline break? Existing benchmarks leave it open for three reasons (\cref{tab:comparison}).
(1) \textbf{Stage-isolated evaluation.} Prior benchmarks score either code-generated 3D scenes without rendering~\cite{wang2026cinerecon,diao2026vibevideobench} or video generation models fed error-free engine recordings~\cite{huang2026prowbench,qiu2026renderbench,liu2026anygamerenderer}, so an end-to-end failure cannot be traced to \texttt{T2C} or \texttt{C2V}.
(2) \textbf{Uncontrolled complexity.} Instructions vary many entangled factors at once, so the benefit of clay cannot be measured against compositional complexity~\cite{wu2024conceptmix,yu2024skillmix}.
(3) \textbf{Unexplored clay representations.} Video generation models can condition on coarse blocks, rigged meshes, or depth-semantic maps~\cite{chen2026codeworld}, yet how reliably agents build each from text and how well video generation models render each remain unknown.

\begin{table}[t]
\centering
\caption{\textbf{Comparison with related benchmarks.} \texttt{T2C} and \texttt{C2V} denote Text-to-Clay and Clay-to-Video evaluation, and Real source denotes cases derived from real video shots.}
\label{tab:positioning}\label{tab:comparison}
\vspace{-2pt}
\newcommand{\tY}{\textcolor[HTML]{1A9A3A}{\ding{51}}}
\newcommand{\tN}{\textcolor[HTML]{D62828}{\ding{55}}}
\footnotesize
\setlength{\tabcolsep}{2.5pt}
\resizebox{\linewidth}{!}{%
\begin{tabular}{@{}lcccccc@{}}
\toprule
& \multicolumn{2}{c}{Stage evaluated} & \multirow{2}{*}{\makecell{Multiple\\proxy forms}} & \multirow{2}{*}{\makecell{Controllable\\difficulty}} & \multirow{2}{*}{\makecell{Real\\source}} & \multirow{2}{*}{\makecell{VLM as\\judge}} \\
\cmidrule(lr){2-3}
Work & T2C & C2V & & & & \\
\midrule
\rowcolor[HTML]{EEF3FB}\multicolumn{7}{@{}l}{\textbf{\textit{Coding agents for 3D/4D}}}\\
CineRecon~\cite{wang2026cinerecon} & \tY & \tN & \tY & \tN & \tY & \tY \\
VibeVideoBench~\cite{diao2026vibevideobench} & \tY & \tN & \tN & \tN & \tN & \tY \\
\rowcolor[HTML]{EEF3FB}\multicolumn{7}{@{}l}{\textbf{\textit{Proxy-conditioned video}}}\\
PROWBench~\cite{huang2026prowbench} & \tN & \tY & \tY & \tN & \tN & \tY \\
RenderBench~\cite{qiu2026renderbench} & \tN & \tY & \tN & \tN & \tY & \tN \\
AnyGameRenderer~\cite{liu2026anygamerenderer} & \tN & \tY & \tY & \tN & \tN & \tN \\
Proxy2World~\cite{xu2026proxy2world} & \tN & \tY & \tY & \tN & \tN & \tY \\
OmniVBench~\cite{li2026omnivbench} & \tN & \tY & \tY & \tN & \tY & \tY \\
\midrule
\rowcolor[HTML]{FDF6D8}\textbf{ClayBench (ours)} & \tY & \tY & \tY & \tY & \tY & \tY \\
\bottomrule
\end{tabular}}
\end{table}

\begin{figure*}[t]
\centering
\includegraphics[width=\linewidth,trim=0 141 0 0,clip]{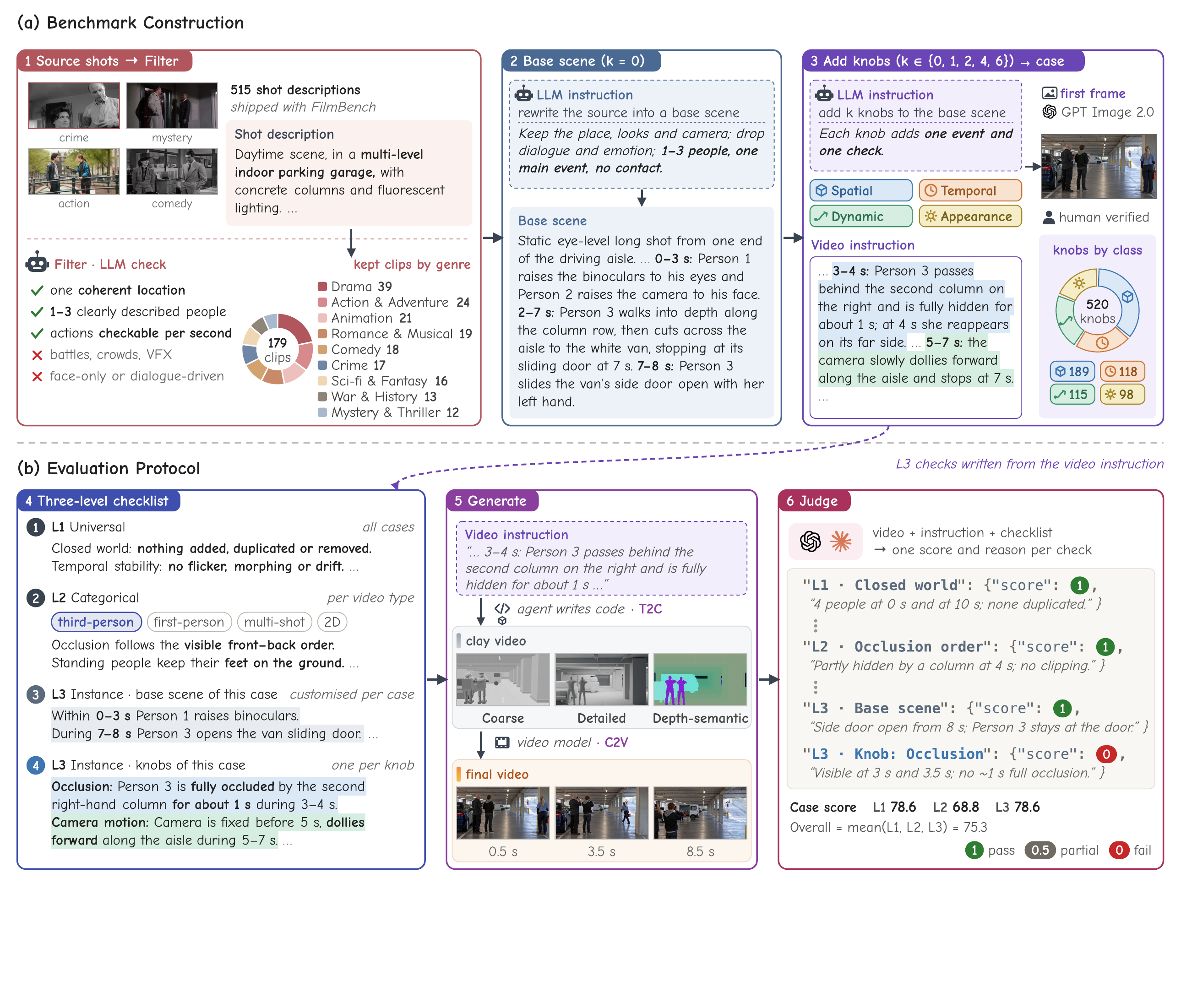}
\caption{\textbf{Construction and evaluation of ClayBench.} (a) Benchmark construction turns real video shots into cases. (b) Evaluation scores both pipeline stages. (1) Shot descriptions are filtered and balanced across genres. (2) Each becomes a single-event base scene. (3) $k$ difficulty knobs and a first frame are added. (4) Each case has a three-level checklist. (5) A coding agent builds the clay video (\texttt{T2C}) and a video generation model renders it (\texttt{C2V}). (6) Two cross-family VLM judges score both videos. All cases are verified by humans.}

\label{fig:pipeline}
\end{figure*}

To bridge these gaps, we introduce \textbf{ClayBench}, a diagnostic benchmark derived from real video shots~\cite{wang2026filmbench} across four video types: third-person, first-person, multi-shot, and 2D animation (\cref{fig:pipeline}). Given a video instruction and a first frame, a coding agent writes code to render a clay video (\texttt{T2C}), which a video generation model renders into the final video (\texttt{C2V}). ClayBench is built around three design principles:
(1) \textbf{Stage-wise diagnostic checklist.} Building on rubric-based VLM evaluation~\cite{li2026omnivbench,zeng2026videoargus}, we score \emph{both} the clay video and the final video with a three-level checklist (\emph{L1 Universal} quality, \emph{L2 Categorical} shooting conventions, and \emph{L3 Instance} spatiotemporal events), attributing every passed or failed constraint to \texttt{T2C} or \texttt{C2V}.
(2) \textbf{Controllable difficulty knobs.} Each case adds $k$ atomic knobs to a single-event base scene, spanning \emph{Spatial}, \emph{Temporal}, \emph{Dynamic}, and \emph{Appearance} constraints, so $k$ controls task difficulty. Comparing against an Image-to-Video (\texttt{I2V}) baseline across $k$ measures how the benefit of clay scales with difficulty.
(3) \textbf{Multi-representation clay comparison.} For every case, coding agents build three clay representations (\emph{Coarse} blocking in Three.js~\cite{cabello2010threejs}, \emph{Detailed} rigged human meshes in Blender~\cite{blender2024}, and \emph{Depth-Semantic} maps exported from the detailed scene), exposing the interplay between coding difficulty and rendering fidelity.

Our evaluation of frontier coding agents and video generation models uncovers three key insights.
First, video generation models lose much of what the clay gets right: 29 to 45\% of the checks that pass in the clay fail in the final video, and after rendering the quality gaps between coding agents shrink (\cref{sec:where_break}).
Second, clay helps at every difficulty level, and its gain persists as instructions get harder: on Seedance 2.0, clay raises L3 scores by 10 to 16 points at every $k$ (\cref{sec:when_help}).
Third, no single clay representation dominates, and a higher clay score does not guarantee a better final video, and video generation models repair many clay errors: Depth-Semantic clay scores lowest after \texttt{T2C}, yet its rendered videos achieve the highest Overall score on every video generation model (\cref{sec:proxy_tradeoff}).

In summary, our contributions are threefold:
\begin{itemize}
    \item \textbf{Diagnostic Benchmark for Code-then-Paint Video Generation.} We formulate the two-stage Text-to-Clay-to-Video evaluation task and construct ClayBench, comprising real-shot-derived cases across four video types with difficulty controlled by $k$ knobs.
    \item \textbf{Stage-wise Attribution Protocol.} We design a unified three-level checklist scored by cross-family VLM judges that evaluates both intermediate clay videos and final videos, disentangling agentic coding failures (\texttt{T2C}) from generative rendering failures (\texttt{C2V}).
    \item \textbf{Extensive Benchmarking and Insights.} Through extensive experiments across coding agents, clay representations, and video generation models, we show that video generation models lose much of what the clay gets right and that clay helps at every difficulty level, and compare the trade-offs between representations.
\end{itemize}

\section{Related Work}
\label{sec:related}
\begin{figure*}[t]
\centering
\includegraphics[width=\linewidth]{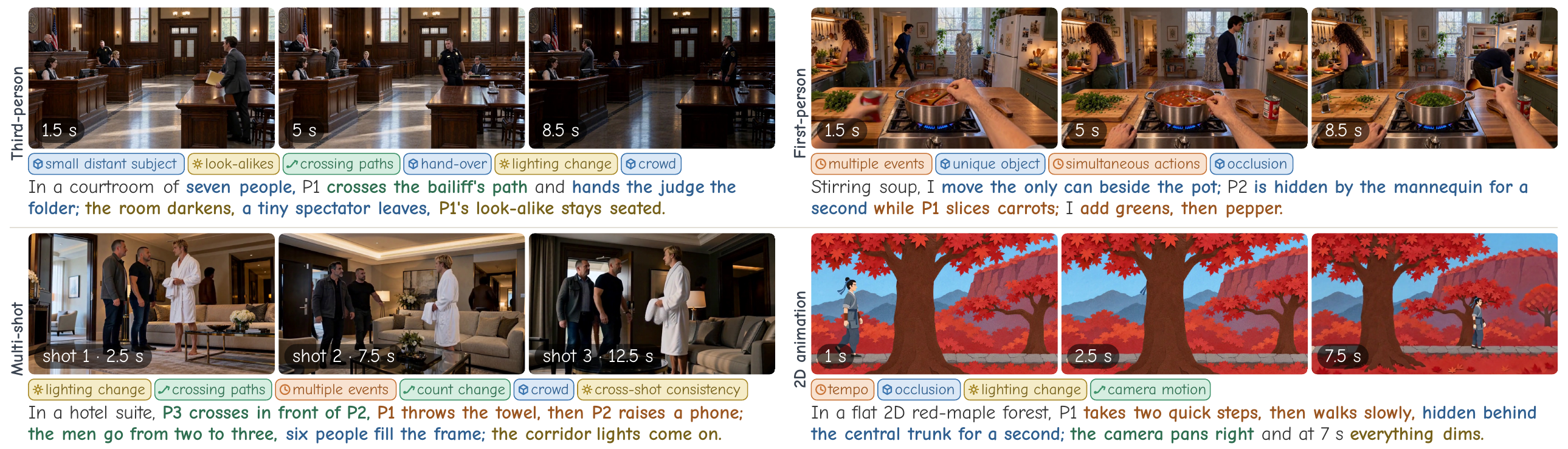}
\caption{\textbf{Qualitative examples from ClayBench}, one harder case per video type, labeled at the left. Each case shows three timestamped frames, its active knobs, and an excerpt of the video instruction. Frames are from MiniMax-H3 videos of Claude Opus 5.5 clays.}

\label{fig:showcase}
\end{figure*}

\subsection{Controllable Video Generation}
\label{sec:rel_control}
Controllable video generation~\cite{ma2025controllable} augments text prompts with explicit conditions to specify scene structure and motion, including depth maps, composable spatiotemporal cues, and camera or object trajectories~\cite{bai2025recammaster,zhang2026flexcomposer,wang2023videocomposer,wang2024motionctrl,kong2026composing,chang2026give}. ControlNet provides a foundational framework for injecting structural conditions into diffusion models~\cite{zhang2023controlnet}, with subsequent work adapting this conditioning strategy to video synthesis~\cite{zhang2024controlvideo,chen2023control,wang2024easycontrol,lin2025ctrl}. With structural conditions serving as an interface to the generator, language-guided planners can automate their specification through dynamic layouts and cross-scene entity plans~\cite{lin2024videodirectorgpt}.

Recent approaches use executable code to construct scenes or maintain evolving world states, and condition visual generation on the resulting structural representations~\cite{lv2024gpt4motion,chen2026codeworld,huang2026pwm,fang2026coder}. Proxy-conditioned renderers translate coarse simulations or other structural representations into detailed videos~\cite{gomez2026coarse,fang2026solarrenderer, xu2026proxy2world}. Together, these advances motivate code-then-paint pipelines, in which explicit proxies carry structure and generators supply appearance. Their intermediate representations vary in geometric detail, motion specificity, and construction cost, raising a practical question about which proxy representations best support faithful generation. We study their reliability when proxies are themselves generated by coding agents, considering both the quality of the constructed proxy and the fidelity of its subsequent rendering.

\subsection{Video Generation Benchmarks}
\label{sec:rel_bench}
Video benchmarks evaluate visual quality, temporal consistency, compositional adherence, and intrinsic faithfulness~\cite{zheng2025vbench2,huang2024vbench,sun2025t2vcompbench,he2026entitybench,shi2026msvbench}. Specialized benchmarks additionally assess physical plausibility and interactions~\cite{bansal2025videophy,meng2024towards}. ConceptMix and Skill-Mix vary the number of compositional requirements~\cite{wu2024conceptmix,yu2024skillmix}, while recent rubric-based evaluation provides instance-specific criteria and diagnostic evidence~\cite{zeng2026videoargus,xu2026vgif,jia2026raven}. These developments inform our difficulty knobs and shared checklist, which target the realization of explicit spatial relations, event timing, and scene dynamics across intermediate and final outputs.

Closer to our setting, agent-side benchmarks assess scene construction and reconstruction, executable dynamic programs, and code-driven video creation~\cite{wang2026cinerecon,shen20264dcodebench,fang2026blenderworld,diao2026vibevideobench,lu2026worldcoderbench}. On the rendering side, PROWBench, RenderBench, ProxyBench, and AnyGameRenderer evaluate whether videos preserve supplied structures, entities, or program-specified events~\cite{huang2026prowbench,qiu2026renderbench,liu2026anygamerenderer,xu2026proxy2world}. OmniVBench further evaluates the preservation and routing of compositional visual references~\cite{li2026omnivbench}. These settings provide complementary views of construction and rendering (\cref{tab:comparison}). ClayBench connects them by evaluating agent-built clay and final videos against a three-level checklist of instruction-grounded constraints. Varying the number of constraints and the proxy representation lets us examine which requirements are satisfied during construction, preserved or lost during rendering, and how proxy quality relates to final-video adherence.

\section{The ClayBench Benchmark}
\label{sec:benchmark}
\begin{figure*}[t]
\centering
\includegraphics[width=\linewidth]{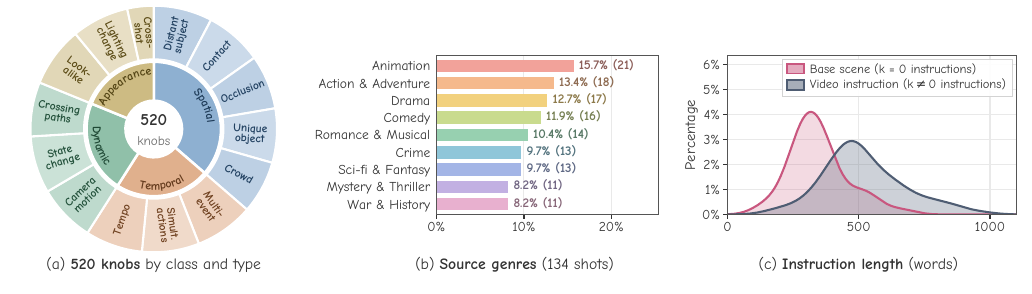}
\caption{\textbf{ClayBench statistics.} (a) 520 knobs of 14 types in four classes, 2.6 per case on average. (b) The 134 source shots span nine genres. (c) Instruction length in words, which grows from 346 for the base scene at $k{=}0$ to 616 at $k{=}6$ on average.}
\label{fig:stats}
\end{figure*}

\subsection{Task Definition}
\label{sec:task}
Each case consists of a video instruction $T$ and a first frame $I_0$. $T$ specifies the people, layout, and per-second events of the scene, and $I_0$ fixes their appearance and initial positions and is generated from $T$ by GPT Image 2.0~\cite{openai2026gptimage2}. The Text-to-Clay-to-Video pipeline splits text-to-video generation into two stages: T2C turns $(T, I_0)$ into a clay video $C$, and C2V turns $(C, I_0, T)$ into the final video $V$. Both videos are scored with a three-level checklist (\cref{sec:eval}), so the two stages can be evaluated separately or end to end.

\paragraph{Clay representations.} We consider three clay representations that differ in geometric detail. \textbf{Coarse} clay represents people and objects with a few geometric primitives in Three.js~\cite{cabello2010threejs}, preserving subject counts, coarse proportions, and the moving parts needed for each action. \textbf{Detailed} clay is a Blender~\cite{blender2024} scene with procedurally modeled layout and props, in which each person is a rigged parametric body from MPFB2~\cite{mpfb2} driven by motion capture. \textbf{Depth-Semantic} clay is exported automatically from the detailed scene as a depth map and a semantic map with 12 classes (such as sky, building, human, and prop), following the proxy format of Code World Model~\cite{chen2026codeworld}.

\paragraph{Text-to-Clay.} A coding agent reads $T$ and $I_0$, writes an executable program, and renders a clay video $C$ that carries geometry only. A clay whose program still fails after one retry scores 0 at both stages.

\paragraph{Clay-to-Video.} A video generation model produces $V$ from $C$, $I_0$, and $T$. The three-level checklist scores $C$ and $V$ separately, so each failure is located in \texttt{T2C} or \texttt{C2V}.

\paragraph{Image-to-Video baseline.} As a baseline, the same video generation model produces $V_{\text{I2V}}$ from $I_0$ and $T$ alone, with only the sentences that refer to the clay removed. The gap between $V$ and $V_{\text{I2V}}$ measures the benefit of clay (\cref{sec:clay_vs_noclay}).

\subsection{Data Construction Pipeline}
\label{sec:construction}
\paragraph{Source data.} We build ClayBench from FilmBench~\cite{wang2026filmbench}, which provides text descriptions of real video shots, in three steps (\cref{fig:pipeline}). \emph{First}, an LLM filters the descriptions and keeps shots whose layout and actions can be specified and checked second by second. \emph{Second}, from the kept shots, we select cases that balance video types and difficulty levels and check each one manually, which yields 134 shots across 9 genres. \emph{Third}, an LLM rewrites each selected description into a base scene, the simplest version of the shot: it keeps the location, the appearance of the people and the camera, removes dialogue and emotions, and specifies each person's position and per-second actions around a single main event.

\paragraph{Difficulty knobs.} A difficulty knob is a controllable source of difficulty: a single requirement, such as an occlusion, a camera move, or a second simultaneous action, that is added to the base scene and makes it harder. The 14 knobs fall into four classes: \emph{Spatial} (occlusion, contact, unique object placement, small distant subject, crowd), \emph{Temporal} (multiple events, simultaneous actions, tempo), \emph{Dynamic} (crossing paths, camera motion, count or state change), and \emph{Appearance} (lighting change, look-alike people, cross-shot consistency). The difficulty $k$ of a case is its number of knobs, $k \in \{0,1,2,4,6\}$, where $k{=}0$ is the base scene. Each knob is verified by one instance check (\cref{sec:eval}).

\paragraph{Case generation.} Starting from the base scene, the LLM writes $k$ knobs into the instruction, and the knobs are assigned so that each one appears evenly across values of $k$. An image model then generates the first frame $I_0$ from the instruction (\cref{fig:pipeline}). Each case has 2.6 knobs on average (\cref{sec:supp_writing}).

\subsection{Dataset Statistics}
\label{sec:stats}
ClayBench contains 200 cases from 134 source shots across 9 genres. By shooting style, the cases fall into four video types of 50 cases each. \emph{First-person} shots are captured from the viewer's eyes with the viewer's hands in view. In \emph{third-person} shots, the camera moves independently of the characters. \emph{Multi-shot} cases cut between shots and require consistent identities and screen directions across cuts. \emph{2D animations} express relations through layering and timing on a flat plane. Each difficulty level $k \in \{0,1,2,4,6\}$ contains 40 cases, and every knob appears in 18--41 cases (\cref{fig:stats}). Higher difficulty yields longer and more crowded instructions: from $k{=}0$ to $k{=}6$, the mean instruction length grows from 346 to 616 words and the mean number of people from 1.8 to 4.6. In total, ClayBench contains 1,125 instance checks, 605 for base scenes and 520 for knobs, with 5.6 per case on average, and 71\% of their time windows span at most 3\,s. \Cref{fig:showcase} shows examples.

\begin{table*}[t]
\centering
\caption{\textbf{Main results on ClayBench.} Scores (\%) of the clay (Clay) and final (Render) videos with Claude Opus 5.5 clays. L3 classes: \textcolor{cSpa}{Spatial (Spa.)}, \textcolor{cTem}{Temporal (Tem.)}, \textcolor{cDyn}{Dynamic (Dyn.)}, \textcolor{cApp}{Appearance (App.)}. Overall: mean of L1--L3. \textbf{Bold}: best per model.}
\label{tab:main}\label{tab:main_results}
\vspace{-2pt}
{\small\setlength{\tabcolsep}{5.35pt}\setlength{\aboverulesep}{0pt}\setlength{\belowrulesep}{0pt}\renewcommand{\arraystretch}{1.04}%
\sbox{\tabbox}{\begin{tabular}{l l | cccc | ccccccccc}
\toprule
& & \multicolumn{4}{>{\columncolor{peach}}c|}{Clay} & \multicolumn{9}{>{\columncolor{ylw}}c}{Render} \\
 &  &  &  &  &  &  &  &  & \multicolumn{5}{c}{\footnotesize L3 by category} &  \\ \cmidrule(lr){10-14}
Video & Clay input & L1 & L2 & L3 & Overall & L1 & L2 & L3 & \catbar{cBase}{Base} & \catbar{cSpa}{Spa.} & \catbar{cTem}{Tem.} & \catbar{cDyn}{Dyn.} & \catbar{cApp}{App.} & Overall \\
\midrule
 \multirow{4}{*}{\makecell[c]{\lgMM\\MiniMax-H3}} & I2V & -- & -- & -- & -- & 83.0 & 82.4 & 66.3 & 72.2 & 58.0 & 56.2 & 44.8 & 71.9 & 77.2 \\
  & Coarse & \textbf{83.2} & \textbf{91.8} & \textbf{79.9} & \textbf{85.0} & 79.5 & 83.8 & 70.8 & 76.7 & \textbf{62.0} & 59.8 & 54.5 & \textbf{72.0} & 78.0 \\
  & Detailed & 82.6 & 91.6 & 78.2 & 84.1 & 80.4 & 83.9 & 70.9 & \textbf{77.2} & 59.9 & 57.6 & \textbf{55.2} & 70.7 & 78.4 \\
  & Depth-Sem. & 79.9 & 90.0 & 70.3 & 80.1 & \textbf{81.5} & \textbf{84.5} & \textbf{71.0} & 77.0 & \textbf{62.0} & \textbf{60.9} & 52.7 & 70.8 & \textbf{79.0} \\
\midrule
 \multirow{4}{*}{\makecell[c]{\lgSD\\Seedance 2.0}} & I2V & -- & -- & -- & -- & 83.6 & 84.0 & 66.7 & 69.4 & 63.0 & 56.2 & 55.9 & 73.1 & 78.1 \\
  & Coarse & \textbf{83.2} & \textbf{91.8} & \textbf{79.9} & \textbf{85.0} & 83.0 & 88.7 & 79.8 & 82.3 & 76.6 & \textbf{72.4} & 69.4 & 83.8 & 83.8 \\
  & Detailed & 82.6 & 91.6 & 78.2 & 84.1 & 84.1 & 88.8 & \textbf{80.8} & \textbf{83.6} & \textbf{77.2} & 69.2 & \textbf{69.8} & \textbf{85.2} & 84.6 \\
  & Depth-Sem. & 79.9 & 90.0 & 70.3 & 80.1 & \textbf{85.6} & \textbf{90.8} & 78.8 & 81.8 & 76.2 & 72.0 & 67.1 & 77.5 & \textbf{85.1} \\
\midrule
 \multirow{4}{*}{\makecell[c]{\lgKL\\Kling 4.0}} & I2V & -- & -- & -- & -- & 85.0 & 84.8 & 69.2 & 73.3 & 66.2 & 54.2 & 54.1 & 72.0 & 79.7 \\
  & Coarse & \textbf{83.2} & \textbf{91.8} & \textbf{79.9} & \textbf{85.0} & 82.0 & 86.6 & 77.2 & 80.5 & 70.0 & 70.9 & 65.8 & \textbf{79.6} & 81.9 \\
  & Detailed & 82.6 & 91.6 & 78.2 & 84.1 & 82.1 & 86.6 & \textbf{77.9} & \textbf{82.3} & \textbf{71.3} & \textbf{72.0} & 67.0 & 73.1 & 82.2 \\
  & Depth-Sem. & 79.9 & 90.0 & 70.3 & 80.1 & \textbf{83.2} & \textbf{88.0} & 77.1 & 81.2 & 70.5 & 67.5 & \textbf{69.2} & 71.8 & \textbf{82.8} \\
\bottomrule
\end{tabular}}%
\typeout{TABW=\the\wd\tabbox}\ifdim\wd\tabbox>\linewidth\resizebox{\linewidth}{!}{\usebox{\tabbox}}\else\usebox{\tabbox}\fi}
\end{table*}

\subsection{Evaluation Methodology}
\label{sec:eval}\label{sec:eval_method}

\paragraph{Three-level checklist.} We decompose each case into atomic checks at three levels, each about one visible fact.

\paragraph{L1 Universal.} Seven checks are shared by all cases: instruction adherence, following the clay, a closed world (no people or props added or missing), identity stability, physical plausibility, visual clarity, and finished appearance.

\paragraph{L2 Categorical.} Four checks target each video type. \emph{Third-person}: camera motion, depth ordering, body support, and background stability. \emph{First-person}: viewpoint, hand-tool continuity, head motion, and manipulated objects. \emph{Multi-shot}: cut timing, cross-shot identity, spatial layout, and action continuity. \emph{2D animation}: flat style, text and symbols, shape consistency, and the order of state changes.

\paragraph{L3 Instance.} Instance checks target the specific case, and each names the people or props involved, a time window, and the expected state. Base-scene checks verify the opening layout, the main event, and the final state, and knob checks verify that each active knob is realized. L1 and L2 are fixed templates, L3 checks are drafted by the LLM with the instruction, and all checks are reviewed manually.

\paragraph{Checklist for clay video.} We also score the clay video $C$ with the three-level checklist. Checks scored on both the clay and the final video locate each failure in T2C or C2V (\cref{sec:attribution}).

\paragraph{VLM as judge.} Since the coding agents and the judges come from the same model families, a single judge may favor outputs of its own family, a tendency known as self-preference bias~\cite{panickssery2024llm,zheng2023judging}. We therefore use two VLM judges from different families, GPT-6.1 Sol and Claude Sonnet 5.5, and average their scores. Each judge receives the video, the instruction used to generate it, and the checklist, without knowing the coding agent or the clay representation, and assigns each check 1 (satisfied), 0.5 (partially satisfied) or 0 (not satisfied). The score of a level is the mean of its checks $\times 100$, and Overall is the mean of L1, L2, and L3.

\section{Experiments}
\label{sec:experiments}

\subsection{Experimental Setup}
\label{sec:exp_setup}
\label{sec:setup}

\noindent\textbf{Coding agents (\texttt{T2C}).}
We evaluate Claude Opus 5.5~\cite{anthropic2026opus55} and Claude Sonnet 5.5~\cite{anthropic2026sonnet55} in Claude Code~\cite{anthropic2026claudecode} and GPT-6.1 Sol~\cite{openai2026sol} in Codex~\cite{openai2026codex}. All agents receive the same structured scene plan, generated once from $T$ and $I_0$ by GPT-6.1 Sol, listing initial positions and event timestamps, and build all three clay representations (\cref{sec:task}).

\noindent\textbf{Video generation models (\texttt{C2V}).}
We evaluate the open-weight MiniMax-H3~\cite{minimax2026h3} ($1344\times768$, 24 fps) and Seedance 2.0~\cite{seedance2026} and Kling 4.0~\cite{kuaishou2026kling4} through their APIs (720p). Each model takes the clay video $C$ as structural reference, the first frame $I_0$ as appearance reference, and the instruction $T$ as prompt.

\noindent\textbf{Image-to-Video (\texttt{I2V}) baseline and evaluation.}
The \texttt{I2V} baseline generates videos from $I_0$ and $T$ alone. Two VLM judges from different families, GPT-6.1 Sol and Claude Sonnet 5.5, score all videos, and we report their mean.

\subsection{Results}\label{sec:results}\label{sec:where_break}\label{sec:attribution}\label{sec:proxy_tradeoff}\label{sec:representations}\label{sec:clay_types}\label{sec:when_help}\label{sec:main_results}\label{sec:difficulty}\label{sec:clay_vs_noclay}
\Cref{tab:main_results} reports end-to-end results with Claude Opus 5.5 clays, and we highlight the key takeaways below.

\noindent\textbf{\texttt{C2V} outperforms \texttt{I2V} on every video generation model, and closed-source models gain more.} Every clay representation raises L2, L3, and Overall over \texttt{I2V} on all three video generation models (\cref{tab:main_results}). The gain is largest on Seedance 2.0, where the best clay adds 14.1 L3 points, and smallest on the open-weight MiniMax-H3, with Kling 4.0 in between at 8.7 points. Under \texttt{I2V}, the three video generation models score within 3 L3 points of one another, so \texttt{I2V} evaluation alone would rank them as nearly equal. With clay they separate clearly, and Seedance 2.0 leads MiniMax-H3 by 9.8 L3 points. Following a clay video is thus a distinct capability, and ClayBench measures it.

\noindent\textbf{Depth-Semantic clay raises Overall at no extra coding cost.} It is exported from the Detailed scene as per-frame depth and semantic maps without another agent call, yet it gives the highest Overall on all three video generation models, 0.6 points above Detailed clay, mainly through higher L1 and L2 scores, while its L3 score on Seedance 2.0 is 2.0 points lower (\cref{tab:main_results}). It does so with the lowest \texttt{T2C} score of the three representations, so a higher clay score does not guarantee a better final video.

\begin{table}[t]
\centering
\caption{\textbf{L3 by video type.} Final-video L3 scores (\%) by video type (Claude Opus 5.5 clays). Overall is the mean of L1, L2, and L3. \textbf{Bold}: best per video generation model.}
\label{tab:by_type}
\vspace{-2pt}
{\small\setlength{\tabcolsep}{2.0pt}\setlength{\aboverulesep}{0pt}\setlength{\belowrulesep}{0pt}\renewcommand{\arraystretch}{1.18}%
\sbox{\tabbox}{\begin{tabular}{l l | c c c c c c}
\toprule
Video & Clay input & 3rd-person & 1st-person & Multi-shot & 2D & L3 All & Overall \\
\midrule
\multirow{4}{*}{\makecell[c]{\lgMM\\MiniMax-H3}} & I2V & 64.4 & 62.8 & 72.7 & 65.2 & 66.3 & 77.2 \\
 & Coarse & \textbf{72.1} & 67.3 & 71.8 & 72.2 & 70.8 & 78.0 \\
 & Detailed & 71.1 & \textbf{67.8} & 72.2 & 72.5 & 70.9 & 78.4 \\
 & Depth-Sem. & 70.8 & 67.5 & \textbf{72.8} & \textbf{72.8} & \textbf{71.0} & \textbf{79.0} \\
\midrule
\multirow{4}{*}{\makecell[c]{\lgSD\\Seedance 2.0}} & I2V & 68.2 & 66.4 & 67.1 & 64.8 & 66.7 & 78.1 \\
 & Coarse & 81.9 & 78.1 & 77.8 & 81.7 & 79.8 & 83.8 \\
 & Detailed & \textbf{83.5} & \textbf{79.2} & \textbf{80.2} & 80.3 & \textbf{80.8} & 84.6 \\
 & Depth-Sem. & 79.8 & 77.2 & 75.6 & \textbf{82.7} & 78.8 & \textbf{85.1} \\
\midrule
\multirow{4}{*}{\makecell[c]{\lgKL\\Kling 4.0}} & I2V & 68.3 & 67.0 & 71.0 & 70.7 & 69.2 & 79.7 \\
 & Coarse & 79.2 & 76.1 & 73.0 & \textbf{80.8} & 77.2 & 81.9 \\
 & Detailed & \textbf{81.0} & \textbf{78.8} & \textbf{74.6} & 77.3 & \textbf{77.9} & 82.2 \\
 & Depth-Sem. & 79.2 & 75.0 & 74.5 & 80.1 & 77.1 & \textbf{82.8} \\
\bottomrule
\end{tabular}}%
\ifdim\wd\tabbox>\linewidth\resizebox{\linewidth}{!}{\usebox{\tabbox}}\else\usebox{\tabbox}\fi}

\end{table}

\noindent\textbf{No single clay is best for every video type.} The overall ranking of representations does not carry over to each video type (\cref{tab:by_type}). Depth-Semantic clay leads L3 in only 3 of the 12 combinations of video generation model and video type. On the two closed-source models, Detailed clay is the stronger choice for third-person, first-person, and multi-shot videos. On MiniMax-H3, the three clays perform almost the same on every video type. The best representation thus depends on both the video generation model and the kind of video to generate.

\begin{figure}[t]
\centering
\includegraphics[width=\linewidth]{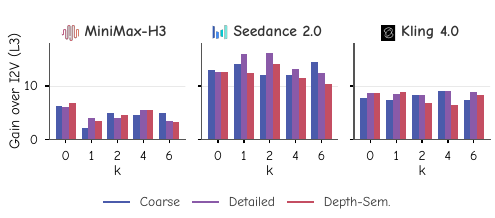}
\caption{\textbf{Gain of clay over \texttt{I2V} by knob count $k$.} L3 gain of each clay over \texttt{I2V} at each $k$ (Claude Opus 5.5 clays).}
\label{fig:by_k}\label{tab:by_k}
\end{figure}

\noindent\textbf{\texttt{C2V} outperforms \texttt{I2V} at every $k$.} Clay leads \texttt{I2V} in all 45 combinations of video generation model, clay representation, and $k$ (\cref{fig:by_k}). As $k$ grows, instructions become harder and absolute scores trend downward for every input, yet the gap between clay and \texttt{I2V} stays roughly constant. Its size depends mainly on the video generation model: 2 to 7 points on MiniMax-H3, 7 to 9 on Kling 4.0, and 10 to 16 on Seedance 2.0. Clay therefore remains useful even for the hardest instructions in ClayBench.

\begin{figure}[t]
\centering
\includegraphics[width=\linewidth]{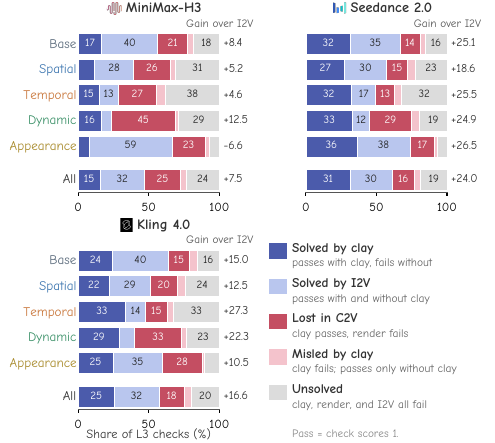}
\caption{\textbf{Stage-wise attribution of L3 checks (\%).} Each check is assigned to one outcome from its clay, final, and I2V verdicts (Claude Opus 5.5 clays). Gain over I2V: share of checks the final video passes minus the share I2V passes.}
\label{fig:attribution}\label{tab:attribution}
\end{figure}

\Cref{fig:attribution} assigns every L3 check to one of five outcomes from the verdicts of the clay video $C$, the final video $V$, and the baseline $V_{\text{I2V}}$, where a check passes only when it scores 1: \oc{oHelp}{Solved by clay} (passes with clay, fails without), \oc{oI2V}{Solved by I2V} (passes with and without clay), \oc{oLost}{Lost in C2V} (clay passes, render fails), \oc{oMis}{Misled by clay} (clay fails, passes only without clay), and \oc{oUns}{Unsolved} (clay, render, and I2V all fail). This attribution separates errors made in \texttt{T2C} from errors made in \texttt{C2V} for every check.

\begin{table}[t]
\centering
\caption{\textbf{Comparison of coding agents (50 cases).} Clay and MiniMax-H3 render scores (\%) across three coding agents. \textbf{Bold}: column best.}
\label{tab:agents}
\vspace{-2pt}
{\small\setlength{\tabcolsep}{6.0pt}\setlength{\aboverulesep}{0pt}\setlength{\belowrulesep}{0pt}\renewcommand{\arraystretch}{1.12}%
\sbox{\tabbox}{\begin{tabular}{l l | cc | cc}
\toprule
 & & \multicolumn{2}{>{\columncolor{peach}}c|}{Clay} & \multicolumn{2}{>{\columncolor{ylw}}c}{Render} \\
Agent & Clay input & L3 & Overall & L3 & Overall \\
\midrule
\multicolumn{2}{l|}{I2V} & -- & -- & 67.0 & 75.6 \\
\midrule
\multirow{3}{*}{\lgO Codex} & Coarse & 68.1 & 77.2 & 71.5 & 77.5 \\
 & Detailed & 56.5 & 68.7 & 67.2 & 72.4 \\
 & Depth-Sem. & 52.0 & 67.9 & 66.0 & 73.1 \\
\midrule
\multirow{3}{*}{\lgC Opus} & Coarse & 73.6 & 81.5 & 71.1 & 78.0 \\
 & Detailed & \textbf{74.8} & \textbf{83.4} & 72.0 & 78.2 \\
 & Depth-Sem. & 65.1 & 79.4 & \textbf{73.0} & \textbf{79.6} \\
\midrule
\multirow{3}{*}{\lgC Sonnet} & Coarse & 67.4 & 76.3 & 71.3 & 78.1 \\
 & Detailed & 63.8 & 73.9 & 70.0 & 77.9 \\
 & Depth-Sem. & 58.2 & 74.2 & 72.6 & \textbf{79.6} \\
\bottomrule
\end{tabular}}%
\ifdim\wd\tabbox>\linewidth\resizebox{\linewidth}{!}{\usebox{\tabbox}}\else\usebox{\tabbox}\fi}
\end{table}

\noindent\textbf{Video generation models lose much of what the clay gets right.} A large share of the checks that pass in the clay fail in the final video, from 29\% on Seedance 2.0 to 45\% on MiniMax-H3 (\cref{fig:attribution}). This loss is far more common than the opposite error, where a wrong clay leads the video generation model away from a check that \texttt{I2V} would pass. Among the checks the final video fails, 41 to 47\% already pass in the clay, and most of the rest come from clay that is only partly correct.

\noindent\textbf{Imperfect clay still yields correct videos.} Video generation models also repair clay errors. Of the checks that fail in the clay, 39 to 49\% pass in the final video. The ranking mirrors the loss: Seedance 2.0 repairs the most (49\%) and loses the fewest clay passes (29\%), whereas MiniMax-H3 repairs the fewest (39\%) and loses the most (45\%). Repair is also more frequent when the clay is partly correct, for 39 to 50\% of the checks scored 0.5 in the clay against 29 to 41\% of those scored 0. Repair is most frequent for Depth-Semantic clay, which has the lowest clay pass rate (43\% vs.\ 57 to 61\%) but the highest repair rate on every video generation model. This is consistent with its top Overall in \cref{tab:main_results}. Because rendering both loses and repairs checks, the clay score and the final score do not move one to one.

\noindent\textbf{Motion and timing need clay, and their remaining errors sit in different stages.} Under \texttt{I2V}, Dynamic and Temporal checks have the lowest pass rates of the four categories on every video generation model (\cref{fig:attribution}). On the two closed-source models, clay raises both by 22 to 27 points. Dynamic checks, which cover how subjects and the camera move, usually pass in the clay, yet \texttt{C2V} loses them more often than any other category. Temporal checks, which cover when events happen and in what order, have the lowest clay pass rate and the highest \emph{Unsolved} share, so their headroom lies in \texttt{T2C}. Appearance checks have the highest \texttt{I2V} pass rate. At the knob level, occlusion is lost most often on every video generation model, in 61 to 86\% of the occlusion checks that pass in the clay, followed on average by crossing paths and camera motion. Crowd checks are lost least, in 9 to 15\%.

\noindent\textbf{Coding agents differ in \texttt{T2C} but converge after rendering.} The three coding agents produce clays of clearly different quality (\cref{tab:agents}). Claude Opus 5.5 leads \texttt{T2C} on all three representations, and GPT-6.1 Sol in Codex struggles with the Detailed and Depth-Semantic formats, where its final Overall even falls below \texttt{I2V}. After rendering on MiniMax-H3, however, the best final L3 of the three agents lies within 1.5 points, while their best clay L3 spans 7.4 points. Opus is also the only agent whose Detailed clay outscores its Coarse clay at both stages. Claude Sonnet 5.5 with Depth-Semantic clay matches Opus for the top final Overall, so the strongest coding agent is not required for the best final video.

\begin{table}[t]
\centering
\caption{\textbf{Agreement between VLM judges and human ratings.}}
\label{tab:human}
\vspace{-2pt}
{\small\setlength{\tabcolsep}{4.98pt}\setlength{\aboverulesep}{0pt}\setlength{\belowrulesep}{0pt}\renewcommand{\arraystretch}{1.12}%
\newcommand{\hp}{\textcolor{red}{--}}%
\sbox{\tabbox}{\begin{tabular}{c | cc | cc}
\toprule
 & \multicolumn{2}{c|}{Clay video} & \multicolumn{2}{c}{Final video} \\
Level & Pearson & Spearman & Pearson & Spearman \\
\midrule
L1 & 0.81 & 0.87 & 0.76 & 0.82 \\
L2 & 0.82 & 0.80 & 0.89 & 0.90 \\
L3 & 0.89 & 0.86 & 0.90 & 0.80 \\
\midrule
Overall & 0.88 & 0.86 & 0.89 & 0.88 \\
\midrule
Human--Human & \multicolumn{2}{c|}{0.79} & \multicolumn{2}{c}{0.76} \\
\bottomrule
\end{tabular}}\usebox{\tabbox}}
\end{table}

\subsection{Human Evaluation}
\label{sec:judge_reliability}
\label{sec:judges}

Three annotators, blind to the video generation model, clay type, and judge scores, rated the clay and final videos of 500 cases on each checklist level, and the two-judge mean correlates strongly with their mean rating at both stages (\cref{tab:human})~\cite{spearman1904association,krippendorff2004reliability}. \Cref{sec:supp_human} gives the protocol and consent.

\section{Conclusion}
\label{sec:conclusion}
We presented ClayBench, a diagnostic benchmark derived from real video shots to evaluate the two-stage Text-to-Clay-to-Video pipeline. By pairing controllable difficulty knobs ($k$) with a unified three-level checklist scored on both clay videos and final videos, ClayBench disentangles coding failures (\texttt{T2C}) from rendering failures (\texttt{C2V}). Our evaluation across three coding agents, three clay representations, and three video generation models shows that clay conditioning surpasses \texttt{I2V} at every difficulty level, yet video generation models lose many of the checks that pass in the clay and compress the gaps between coding agents. The two stages fail on different constraints: temporal sequencing is limited mainly by \texttt{T2C}, whereas dynamic motion is lost mainly in \texttt{C2V}. By locating each failure in \texttt{T2C} or \texttt{C2V}, ClayBench shows where progress pays off most: temporal sequencing in coding agents and motion fidelity in video generation models.

\section*{LLM Usage Statement}
LLMs are part of the studied pipeline: coding agents (Claude Opus 5.5 and Claude Sonnet 5.5 in Claude Code, and GPT-6.1 Sol in Codex) build the clay videos, GPT Image 2.0 generates the first frames, and GPT-6.1 Sol and Claude Sonnet 5.5 serve as VLM judges. LLMs also assisted in drafting video instructions and checklists, which the authors reviewed and revised, and in polishing the writing. The authors take full responsibility for all content.

{
    \small
    \bibliographystyle{ieeenat_fullname}
    \bibliography{main}

\begin{thebibliography}{58}
\providecommand{\natexlab}[1]{#1}
\providecommand{\url}[1]{\texttt{#1}}
\expandafter\ifx\csname urlstyle\endcsname\relax
  \providecommand{\doi}[1]{doi: #1}\else
  \providecommand{\doi}{doi: \begingroup \urlstyle{rm}\Url}\fi

\bibitem[{Anthropic}(2026{\natexlab{a}})]{anthropic2026claudecode}
{Anthropic}.
\newblock {Claude Code}: Overview.
\newblock \url{https://code.claude.com/docs/en/overview}, 2026{\natexlab{a}}.
\newblock Living documentation, accessed 2026-11-16.

\bibitem[{Anthropic}(2026{\natexlab{b}})]{anthropic2026opus55}
{Anthropic}.
\newblock {Claude Opus 5.5}.
\newblock \url{https://www.anthropic.com/claude-opus-5-5}, 2026{\natexlab{b}}.

\bibitem[{Anthropic}(2026{\natexlab{c}})]{anthropic2026sonnet55}
{Anthropic}.
\newblock {Claude Sonnet 5.5}.
\newblock \url{https://www.anthropic.com/claude-sonnet-5-5},
  2026{\natexlab{c}}.

\bibitem[Bai et~al.(2025)Bai, Xia, Fu, Wang, Mu, Cao, Liu, Hu, Bai, Wan, and
  Zhang]{bai2025recammaster}
Jianhong Bai, Menghan Xia, Xiao Fu, Xintao Wang, Lianrui Mu, Jinwen Cao, Zuozhu
  Liu, Haoji Hu, Xiang Bai, Pengfei Wan, and Di Zhang.
\newblock {ReCamMaster}: Camera-controlled generative rendering from a single
  video.
\newblock In \emph{2025 IEEE/CVF International Conference on Computer Vision
  (ICCV)}, pages 14834--14844. IEEE, 2025.

\bibitem[Bansal et~al.(2025)Bansal, Lin, Xie, Zong, Yarom, Bitton, Jiang, Sun,
  Chang, and Grover]{bansal2025videophy}
Hritik Bansal, Zongyu Lin, Tianyi Xie, Zeshun Zong, Michal Yarom, Yonatan
  Bitton, Chenfanfu Jiang, Yizhou Sun, Kai-Wei Chang, and Aditya Grover.
\newblock {VideoPhy}: Evaluating physical commonsense for video generation.
\newblock In \emph{ICLR}, 2025.

\bibitem[{Blender Foundation}(2026)]{blender2024}
{Blender Foundation}.
\newblock Blender: Free and open source {3D} creation software.
\newblock \url{https://github.com/blender/blender}, 2026.
\newblock Software documentation, accessed 2026-11-16.

\bibitem[Cabello et~al.(2026)]{cabello2010threejs}
Ricardo Cabello et~al.
\newblock Three.js: {JavaScript} {3D} library.
\newblock \url{https://threejs.org}, 2026.
\newblock Software documentation, accessed 2026-11-16.

\bibitem[Chang et~al.(2026)Chang, Lyu, Huang, Tan, Zhao, Ding, Bao, Tao, Wan,
  and Qi]{chang2026give}
Chirui Chang, Xiaoyang Lyu, Yi-Hua Huang, Haoru Tan, Shizhen Zhao, Yikang Ding,
  Jianmin Bao, Xin Tao, Pengfei Wan, and Xiaojuan Qi.
\newblock Geometry-instructed video editing.
\newblock \emph{arXiv preprint arXiv:2606.24225}, 2026.

\bibitem[Chen et~al.(2023)Chen, Ji, Wu, Wu, Xie, Li, Xia, Xiao, and
  Lin]{chen2023control}
Weifeng Chen, Yatai Ji, Jie Wu, Hefeng Wu, Pan Xie, Jiashi Li, Xin Xia, Xuefeng
  Xiao, and Liang Lin.
\newblock {Control-A-Video: Controllable Text-to-Video Diffusion Models with
  Motion Prior and Reward Feedback Learning}.
\newblock \emph{arXiv preprint arXiv:2305.13840}, 2023.

\bibitem[Chen et~al.(2026)Chen, Lin, and Zhang]{chen2026codeworld}
Yiwen Chen, Guosheng Lin, and Chi Zhang.
\newblock Code world model: Coding agent as world brain.
\newblock \emph{arXiv preprint arXiv:2608.25927}, 2026.

\bibitem[Diao et~al.(2026)Diao, Xu, Yang, Ma, Chen, Chen, Jiang, Liu, Zhang,
  Fu, Wang, Zhang, Fan, Li, Li, Wang, Zheng, Wen, Gao, Li, Zhang, Xu, Liang,
  and Ma]{diao2026vibevideobench}
Muxi Diao, Wenbin Xu, Lele Yang, Xuyang Ma, Yuzhe Chen, Jiankang Chen, Kaiyu
  Jiang, Changyi Liu, Haoyu Zhang, Hongyi Fu, Jianxiong Wang, Tianke Zhang,
  Haonan Fan, Yingxin Li, Bingyi Li, Xu Wang, Xuanyu Zheng, Bin Wen, Tingting
  Gao, Han Li, Wentao Zhang, Weiran Xu, Kongming Liang, and Zhanyu Ma.
\newblock {VibeVideoBench}: Evaluating multimodal agents on code-driven video
  generation.
\newblock \emph{arXiv preprint arXiv:2610.35864}, 2026.

\bibitem[Fang et~al.(2026{\natexlab{a}})Fang, Qian, Lu, Huang, Jiang, Wang,
  Song, Jiang, and Shou]{fang2026solarrenderer}
Guian Fang, Shengju Qian, Dongyue Lu, Junchao Huang, Lutao Jiang, Xin Wang,
  Yiren Song, Li Jiang, and Mike~Zheng Shou.
\newblock {SolarRenderer: Few-Step Generative Reference Rendering}.
\newblock \emph{arXiv preprint arXiv:2610.28630}, 2026{\natexlab{a}}.

\bibitem[Fang et~al.(2026{\natexlab{b}})Fang, Wang, Yu, Fang, Li, Li, Gu, Wei,
  Zang, Li, Lin, Zhang, Bai, Yang, and Liu]{fang2026blenderworld}
Xinyu Fang, Yuchi Wang, Weihao Yu, Rongyao Fang, Zhaohai Li, Zhixun Li,
  Tiancheng Gu, Xilin Wei, Yuhang Zang, Hongsheng Li, Dahua Lin, Guofeng Zhang,
  Shuai Bai, Zhibo Yang, and Dayiheng Liu.
\newblock {BlenderWorld}: Benchmarking visual agents for {3D} creation from
  real-world references.
\newblock \emph{arXiv preprint arXiv:2610.23986}, 2026{\natexlab{b}}.

\bibitem[Fang et~al.(2026{\natexlab{c}})Fang, Shao, Zhu, Xiao, Chen, Liu, Fu,
  Cao, Zhai, and Zha]{fang2026coder}
Zixun Fang, Yawen Shao, Kai Zhu, Jie Xiao, Shihan Chen, Yu Liu, Xueyang Fu,
  Yang Cao, Wei Zhai, and Zheng-Jun Zha.
\newblock {Code Plans, Diffusion Renders: Open-Ended Generative World
  Modeling}.
\newblock \emph{arXiv preprint arXiv:2609.26458}, 2026{\natexlab{c}}.

\bibitem[Gomez-Nogales et~al.(2026)Gomez-Nogales, Hong, Ge, Zhuang,
  Comino-Trinidad, Casas, and Zhou]{gomez2026coarse}
Gonzalo Gomez-Nogales, Yicong Hong, Chongjian Ge, Peiye Zhuang, Marc
  Comino-Trinidad, Dan Casas, and Yi Zhou.
\newblock Coarse-to-real: Generative rendering for populated dynamic scenes.
\newblock \emph{arXiv preprint arXiv:2601.22301}, 2026.

\bibitem[He et~al.(2026)He, Wei, Yang, and Ordonez]{he2026entitybench}
Ruozhen He, Meng Wei, Ziyan Yang, and Vicente Ordonez.
\newblock {EntityBench}: Towards entity-consistent long-range multi-shot video
  generation.
\newblock \emph{arXiv preprint arXiv:2605.15199}, 2026.

\bibitem[Huang et~al.(2024)Huang, He, Yu, Zhang, Si, Jiang, Zhang, Wu, Jin,
  Chanpaisit, Wang, Chen, Wang, Lin, Qiao, and Liu]{huang2024vbench}
Ziqi Huang, Yinan He, Jiashuo Yu, Fan Zhang, Chenyang Si, Yuming Jiang, Yuanhan
  Zhang, Tianxing Wu, Qingyang Jin, Nattapol Chanpaisit, Yaohui Wang, Xinyuan
  Chen, Limin Wang, Dahua Lin, Yu Qiao, and Ziwei Liu.
\newblock {VBench}: Comprehensive benchmark suite for video generative models.
\newblock In \emph{CVPR}, 2024.

\bibitem[Huang et~al.(2026{\natexlab{a}})Huang, Lin, Lin, Huang, Yu, Niu, Yang,
  Liu, Chuang, Zhang, and Wang]{huang2026pwm}
Zheng-Hui Huang, Guixu Lin, Jiacheng Lin, Yi-Chuan Huang, Ruihan Yu, Muyao Niu,
  Siqi Yang, Yu-Lun Liu, Yung-Yu Chuang, Kaipeng Zhang, and Zhixiang Wang.
\newblock Programmable world model.
\newblock \emph{arXiv preprint arXiv:2609.10540}, 2026{\natexlab{a}}.

\bibitem[Huang et~al.(2026{\natexlab{b}})Huang, Lin, Tsai, Zhu, Lan, Liu,
  Chuang, Zhang, and Wang]{huang2026prowbench}
Zheng-Hui Huang, Guixu Lin, Yu-Ju Tsai, Jian-Kai Zhu, Fengbo Lan, Yu-Lun Liu,
  Yung-Yu Chuang, Kaipeng Zhang, and Zhixiang Wang.
\newblock {PROWBench}: Do video models render what the program specifies?
\newblock \emph{arXiv preprint arXiv:2610.02205}, 2026{\natexlab{b}}.

\bibitem[Jia et~al.(2026)Jia, Qian, Zhang, Zhu, Gao, and Min]{jia2026raven}
Ziheng Jia, Jiaying Qian, Zicheng Zhang, Xiaorong Zhu, Lancheng Gao, and
  Xiongkuo Min.
\newblock {RAVEN-Eval}: Rubric-guided automatic evaluation for {AI} video
  generation models based on {LMM} preference judgement.
\newblock \emph{arXiv preprint arXiv:2608.09111}, 2026.

\bibitem[Kong et~al.(2026)Kong, Zhang, Guo, Zhao, Zhang, and
  Rao]{kong2026composing}
Xianghao Kong, Zeyu Zhang, Yuwei Guo, Zhuoran Zhao, Songchun Zhang, and Anyi
  Rao.
\newblock Composing concepts from images and videos via concept-prompt binding.
\newblock In \emph{CVPR}, pages 14800--14810, 2026.

\bibitem[Krippendorff(2004)]{krippendorff2004reliability}
Klaus Krippendorff.
\newblock Reliability in content analysis: Some common misconceptions and
  recommendations.
\newblock \emph{Human Communication Research}, 30\penalty0 (3):\penalty0
  411--433, 2004.

\bibitem[{Kuaishou Technology}(2026)]{kuaishou2026kling4}
{Kuaishou Technology}.
\newblock Kling {AI} officially launches next-generation {Kling 4.0}, raising
  the bar for professional-grade video models.
\newblock \url{https://kuaishou.gcs-web.com/node/11766/pdf}, 2026.

\bibitem[Li et~al.(2026)Li, Guan, Jiang, Cai, Liu, Zhang, Guan, Huang, Li, Wu,
  Yu, Zhao, Zhao, Liu, Chen, Liu, and Zhu]{li2026omnivbench}
Wenxue Li, Peiyan Guan, Haoyang Jiang, Junxian Cai, Hualuo Liu, Chunjie Zhang,
  Chong Guan, Kai Huang, Songlian Li, Taiyi Wu, Yongjian Yu, Xiaotong Zhao,
  Alan Zhao, Quande Liu, Xi Chen, Yu Liu, and Lei Zhu.
\newblock {OmniVBench}: A benchmark and large-scale dataset for omni
  reference-to-video generation.
\newblock \emph{arXiv preprint arXiv:2609.22069}, 2026.

\bibitem[Lin et~al.(2024)Lin, Zala, Cho, and Bansal]{lin2024videodirectorgpt}
Han Lin, Abhay Zala, Jaemin Cho, and Mohit Bansal.
\newblock {VideoDirectorGPT}: Consistent multi-scene video generation via
  {LLM}-guided planning.
\newblock In \emph{COLM}, 2024.

\bibitem[Lin et~al.(2025)Lin, Cho, Zala, and Bansal]{lin2025ctrl}
Han Lin, Jaemin Cho, Abhay Zala, and Mohit Bansal.
\newblock {Ctrl-Adapter}: An efficient and versatile framework for adapting
  diverse controls to any diffusion model.
\newblock In \emph{International Conference on Learning Representations}, 2025.

\bibitem[Liu et~al.(2026)Liu, Ouyang, Lan, Wang, and
  Zhang]{liu2026anygamerenderer}
Ruicong Liu, Liangyang Ouyang, Fengbo Lan, Zhixiang Wang, and Kaipeng Zhang.
\newblock {AnyGameRenderer}: Entity-consistent generative rendering across
  games and an entity-level benchmark.
\newblock \emph{arXiv preprint arXiv:2610.31466}, 2026.

\bibitem[Lu et~al.(2026)Lu, Xu, Yu, Jiang, Yu, Wang, Yang, Zhang, Wang, He, and
  Liang]{lu2026worldcoderbench}
Shuo Lu, Yinuo Xu, Kecheng Yu, Siru Jiang, Yongcan Yu, Yubin Wang, Haitao Yang,
  Yuxiang Zhang, Bin Wang, Ran He, and Jian Liang.
\newblock {WorldCoder-Bench: Benchmarking Physically Grounded 3D World
  Synthesis}.
\newblock \emph{arXiv preprint arXiv:2606.01869}, 2026.

\bibitem[Lv et~al.(2024)Lv, Huang, Yan, Huang, Liu, Liu, Wen, Chen, and
  Chen]{lv2024gpt4motion}
Jiaxi Lv, Yi Huang, Mingfu Yan, Jiancheng Huang, Jianzhuang Liu, Yifan Liu,
  Yafei Wen, Xiaoxin Chen, and Shifeng Chen.
\newblock {GPT4Motion}: Scripting physical motions in text-to-video generation
  via {Blender}-oriented {GPT} planning.
\newblock In \emph{2024 IEEE/CVF Conference on Computer Vision and Pattern
  Recognition Workshops (CVPRW)}, pages 1430--1440. IEEE, 2024.

\bibitem[Ma et~al.(2025)Ma, Feng, Hu, Wang, Wang, Zheng, Wang, Wang, He, Wang,
  Zhu, Liu, He, Wang, Li, Li, Han, Guo, Liu, Xu, Zhang, and
  Chen]{ma2025controllable}
Yue Ma, Kunyu Feng, Zhongyuan Hu, Xinyu Wang, Yucheng Wang, Mingzhe Zheng,
  Bingyuan Wang, Qinghe Wang, Xuanhua He, Hongfa Wang, Chenyang Zhu, Hongyu
  Liu, Yingqing He, Zeyu Wang, Zhifeng Li, Xiu Li, Sirui Han, Yike Guo, Wei
  Liu, Dan Xu, Linfeng Zhang, and Qifeng Chen.
\newblock {Controllable Video Generation: A Survey}.
\newblock \emph{arXiv preprint arXiv:2507.16869}, 2025.

\bibitem[{MakeHuman Community}(2026)]{mpfb2}
{MakeHuman Community}.
\newblock {MPFB2}: {MakeHuman} plugin for {Blender}.
\newblock \url{https://github.com/makehumancommunity/mpfb2}, 2026.
\newblock Software documentation, accessed 2026-11-16.

\bibitem[Meng et~al.(2025)Meng, Liao, Tan, Lu, Shao, Zhang, Cheng, Li, and
  Luo]{meng2024towards}
Fanqing Meng, Jiaqi Liao, Xinyu Tan, Quanfeng Lu, Wenqi Shao, Kaipeng Zhang, Yu
  Cheng, Dianqi Li, and Ping Luo.
\newblock {Towards World Simulator: Crafting Physical Commonsense-Based
  Benchmark for Video Generation}.
\newblock In \emph{ICML}, pages 43781--43806, 2025.

\bibitem[{MiniMax}(2026)]{minimax2026h3}
{MiniMax}.
\newblock {MiniMax H3}: An open model breaking the boundaries between tasks and
  modalities.
\newblock \url{https://minimax.io/blog/minimax-h3}, 2026.

\bibitem[{OpenAI}(2026{\natexlab{a}})]{openai2026codex}
{OpenAI}.
\newblock Codex.
\newblock \url{https://openai.com/codex/}, 2026{\natexlab{a}}.
\newblock Product documentation, accessed 2026-11-16.

\bibitem[{OpenAI}(2026{\natexlab{b}})]{openai2026gptimage2}
{OpenAI}.
\newblock {GPT-Image-2} model.
\newblock \url{https://developers.openai.com/api/docs/models/gpt-image-2},
  2026{\natexlab{b}}.
\newblock Model documentation, accessed 2026-11-16.

\bibitem[{OpenAI}(2026{\natexlab{c}})]{openai2026sol}
{OpenAI}.
\newblock Introducing {GPT-6.1 Sol}.
\newblock \url{https://openai.com/index/introducing-gpt-6-1-sol/},
  2026{\natexlab{c}}.

\bibitem[Panickssery et~al.(2024)Panickssery, Bowman, and
  Feng]{panickssery2024llm}
Arjun Panickssery, Samuel~R. Bowman, and Shi Feng.
\newblock {LLM} evaluators recognize and favor their own generations.
\newblock In \emph{NeurIPS}, 2024.

\bibitem[Qiu et~al.(2026)Qiu, Xu, Liu, Han, and Ye]{qiu2026renderbench}
Dicong Qiu, Zhiyuan Xu, Yaosheng Liu, Feng Han, and Bo Ye.
\newblock {RenderBench}: Benchmarking render-to-real video transfer with
  reconstructed digital twins.
\newblock \emph{arXiv preprint arXiv:2610.08684}, 2026.

\bibitem[Shen et~al.(2026)Shen, Kova{\v{c}}i{\v{c}}, Kulits, Wang, Li,
  Tenenbaum, Yuille, Chen, and Wu]{shen20264dcodebench}
Ruihong Shen, {\v{Z}}iga Kova{\v{c}}i{\v{c}}, Peter Kulits, Xingrui Wang,
  Zizhang Li, Joshua~B. Tenenbaum, Alan Yuille, Jieneng Chen, and Jiajun Wu.
\newblock {4DCodeBench}: Benchmarking agents on inverse graphics of dynamic
  scenes.
\newblock \emph{arXiv preprint arXiv:2610.03715}, 2026.

\bibitem[Shi et~al.(2026)Shi, Li, Deng, Xu, Chen, Wang, Hu, and
  Zhang]{shi2026msvbench}
Haoyuan Shi, Yunxin Li, Nanhao Deng, Zhenran Xu, Xinyu Chen, Longyue Wang,
  Baotian Hu, and Min Zhang.
\newblock {MSVBench}: Towards human-level evaluation of multi-shot video
  generation.
\newblock In \emph{Findings of the Association for Computational Linguistics:
  ACL 2026}, pages 24034--24058. Association for Computational Linguistics,
  2026.

\bibitem[Spearman(1904)]{spearman1904association}
C. Spearman.
\newblock The proof and measurement of association between two things.
\newblock \emph{The American Journal of Psychology}, 15\penalty0 (1):\penalty0
  72--101, 1904.

\bibitem[Sun et~al.(2025)Sun, Huang, Liu, Wu, Xu, Li, and
  Liu]{sun2025t2vcompbench}
Kaiyue Sun, Kaiyi Huang, Xian Liu, Yue Wu, Zihan Xu, Zhenguo Li, and Xihui Liu.
\newblock {T2V-CompBench}: A comprehensive benchmark for compositional
  text-to-video generation.
\newblock In \emph{CVPR}, 2025.

\bibitem[{Team Seedance} et~al.(2026)]{seedance2026}
{Team Seedance} et~al.
\newblock {Seedance 2.0}: Advancing video generation for world complexity.
\newblock \emph{arXiv preprint arXiv:2604.14148}, 2026.

\bibitem[Wang et~al.(2024{\natexlab{a}})Wang, Gu, Hu, Zhao, Guo, Han, Xu, and
  Liang]{wang2024easycontrol}
Cong Wang, Jiaxi Gu, Panwen Hu, Haoyu Zhao, Yuanfan Guo, Jianhua Han, Hang Xu,
  and Xiaodan Liang.
\newblock {EasyControl: Transfer ControlNet to Video Diffusion for Controllable
  Generation and Interpolation}.
\newblock \emph{arXiv preprint arXiv:2408.13005}, 2024{\natexlab{a}}.

\bibitem[Wang et~al.(2026{\natexlab{a}})Wang, Wang, Wu, Zhang, Chen, Huang,
  Zhang, Pang, and Zhao]{wang2026cinerecon}
Jialei Wang, Zehan Wang, XiaoYang Wu, Ziang Zhang, Siyu Chen, Jiaer Huang,
  Yujia Zhang, Tianyu Pang, and Hengshuang Zhao.
\newblock {CineRecon}: Benchmarking frontier models for agentic cinematic
  reconstruction.
\newblock \emph{arXiv preprint arXiv:2610.16577}, 2026{\natexlab{a}}.

\bibitem[Wang et~al.(2026{\natexlab{b}})Wang, Li, Hu, Qi, Ding, Qiao, Wang, Lv,
  Han, Li, Ding, Wang, Wu, Chen, Wang, Wu, Huang, Zhu, Tian, Li, Fan, Tang, Li,
  Qiang, Chen, Zhen, Zhao, Qu, Li, and Wei]{wang2026filmbench}
Shengyi Wang, Niantong Li, Guangzheng Hu, Hong Qi, Fei Ding, Weixu Qiao, Jinlin
  Wang, Xiaotong Lv, Peng Han, Zimeng Li, Fanshu Ding, Yushu Wang, Han Wu,
  Jingjing Chen, Chongxiao Wang, Yanhao Wu, Chenglong Huang, Xiaoqian Zhu, Jie
  Tian, Hua Li, Jingjing Fan, Mingshuang Tang, Zhong Li, Hengxia Qiang, Weibin
  Chen, Jinyang Zhen, Bing Zhao, Lin Qu, Jing Li, and Hu Wei.
\newblock {FilmBench}: A film-grade benchmark for cinematic video generation.
\newblock \emph{arXiv preprint arXiv:2607.24241}, 2026{\natexlab{b}}.

\bibitem[Wang et~al.(2023)Wang, Yuan, Zhang, Chen, Wang, Zhang, Shen, Zhao, and
  Zhou]{wang2023videocomposer}
Xiang Wang, Hangjie Yuan, Shiwei Zhang, Dayou Chen, Jiuniu Wang, Yingya Zhang,
  Yujun Shen, Deli Zhao, and Jingren Zhou.
\newblock {VideoComposer}: Compositional video synthesis with motion
  controllability.
\newblock In \emph{NeurIPS}, 2023.

\bibitem[Wang et~al.(2024{\natexlab{b}})Wang, Yuan, Wang, Li, Chen, Xia, Luo,
  and Shan]{wang2024motionctrl}
Zhouxia Wang, Ziyang Yuan, Xintao Wang, Yaowei Li, Tianshui Chen, Menghan Xia,
  Ping Luo, and Ying Shan.
\newblock {MotionCtrl}: A unified and flexible motion controller for video
  generation.
\newblock In \emph{SIGGRAPH}, 2024{\natexlab{b}}.

\bibitem[Wu et~al.(2024)Wu, Yu, Huang, Russakovsky, and
  Arora]{wu2024conceptmix}
Xindi Wu, Dingli Yu, Yangsibo Huang, Olga Russakovsky, and Sanjeev Arora.
\newblock {ConceptMix}: A compositional image generation benchmark with
  controllable difficulty.
\newblock In \emph{NeurIPS Datasets and Benchmarks Track}, 2024.

\bibitem[Xu et~al.(2026{\natexlab{a}})Xu, Yan, Wang, Zou, Hong, and
  Huang]{xu2026proxy2world}
Hongli Xu, Weilong Yan, Anbang Wang, Chunyu Zou, Siyu Hong, and Jingwei Huang.
\newblock {Proxy2World}: Learning to generate worlds from lightweight proxies
  without seeing them.
\newblock \emph{arXiv preprint arXiv:2609.35023}, 2026{\natexlab{a}}.

\bibitem[Xu et~al.(2026{\natexlab{b}})Xu, Wang, Chen, Wang, Diao, Zhang, Liang,
  Lin, and Ma]{xu2026vgif}
Songyu Xu, Xin Wang, Qiang Chen, Xinran Wang, Muxi Diao, Yuxuan Zhang, Kongming
  Liang, Rui Lin, and Zhanyu Ma.
\newblock {VGIF-Score}: Interpretable and diagnostic evaluation of
  spatio-temporal instruction following in video generation.
\newblock \emph{arXiv preprint arXiv:2607.13527}, 2026{\natexlab{b}}.

\bibitem[Yu et~al.(2024)Yu, Kaur, Gupta, Brown-Cohen, Goyal, and
  Arora]{yu2024skillmix}
Dingli Yu, Simran Kaur, Arushi Gupta, Jonah Brown-Cohen, Anirudh Goyal, and
  Sanjeev Arora.
\newblock {Skill-Mix}: A flexible and expandable family of evaluations for {AI}
  models.
\newblock In \emph{ICLR}, 2024.

\bibitem[Zeng et~al.(2026)Zeng, Wang, Yu, Hua, and Luo]{zeng2026videoargus}
Ziyun Zeng, Zixuan Wang, Yongsheng Yu, Hang Hua, and Jiebo Luo.
\newblock {VideoArgus}: Agentic rubric-grounded unified evaluation for video
  generation and editing.
\newblock \emph{arXiv preprint arXiv:2608.05485}, 2026.

\bibitem[Zhang et~al.(2023)Zhang, Rao, and Agrawala]{zhang2023controlnet}
Lvmin Zhang, Anyi Rao, and Maneesh Agrawala.
\newblock Adding conditional control to text-to-image diffusion models.
\newblock In \emph{ICCV}, 2023.

\bibitem[Zhang et~al.(2026)Zhang, Guo, Xiong, Liu, Guo, Zhang, and
  Rao]{zhang2026flexcomposer}
Songchun Zhang, Sitong Guo, Xianghao Xiong, Pengwei Liu, Yuwei Guo, Lvmin
  Zhang, and Anyi Rao.
\newblock Flexcomposer: Unified video compositing from images to dynamic
  footage with flexible trajectory control.
\newblock In \emph{European Conference on Computer Vision}, pages 467--486.
  Springer, 2026.

\bibitem[Zhang et~al.(2024)Zhang, Wei, Jiang, Zhang, Zuo, and
  Tian]{zhang2024controlvideo}
Yabo Zhang, Yuxiang Wei, Dongsheng Jiang, Xiaopeng Zhang, Wangmeng Zuo, and Qi
  Tian.
\newblock {ControlVideo}: Training-free controllable text-to-video generation.
\newblock In \emph{International Conference on Learning Representations}, 2024.

\bibitem[Zheng et~al.(2025)Zheng, Huang, Liu, Zou, He, Zhang, Gu, Zhang, He,
  Zheng, Qiao, and Liu]{zheng2025vbench2}
Dian Zheng, Ziqi Huang, Hongbo Liu, Kai Zou, Yinan He, Fan Zhang, Lulu Gu,
  Yuanhan Zhang, Jingwen He, Wei-Shi Zheng, Yu Qiao, and Ziwei Liu.
\newblock {VBench-2.0}: Advancing video generation benchmark suite for
  intrinsic faithfulness.
\newblock \emph{arXiv preprint arXiv:2503.21755}, 2025.

\bibitem[Zheng et~al.(2023)Zheng, Chiang, Sheng, Zhuang, Wu, Zhuang, Lin, Li,
  Li, Xing, Zhang, Gonzalez, and Stoica]{zheng2023judging}
Lianmin Zheng, Wei-Lin Chiang, Ying Sheng, Siyuan Zhuang, Zhanghao Wu, Yonghao
  Zhuang, Zi Lin, Zhuohan Li, Dacheng Li, Eric~P. Xing, Hao Zhang, Joseph~E.
  Gonzalez, and Ion Stoica.
\newblock Judging {LLM}-as-a-judge with {MT-Bench} and {Chatbot Arena}.
\newblock In \emph{NeurIPS Datasets and Benchmarks Track}, 2023.

\end{thebibliography}
}

\clearpage
\appendix
\renewcommand{\thesection}{\Alph{section}}
\counterwithin{table}{section}\counterwithin{figure}{section}
\section*{Appendix}
\noindent The appendix collects the construction details (\cref{sec:supp_bench}), the checklist templates and human evaluation (\cref{sec:supp_check}), the clay representations and prompts (\cref{sec:supp_pipe}), full results by video type (\cref{sec:supp_results}) and qualitative examples (\cref{sec:supp_qual}).

\section{Benchmark Construction}
\label{sec:supp_bench}\label{sec:supp_source}\label{sec:supp_writing}\label{sec:supp_frames}
ClayBench builds on the 515 text-to-video descriptions of FilmBench~\cite{wang2026filmbench}. GPT-6.1 Sol screens them and accepts those that show one coherent location, a base scene with one to three people, actions that can be checked second by second, and room for more people and props. Battles, crowds, special effects, underwater scenes, dialogue and non-human leads are rejected. We read each accepted candidate in full and select 134 source shots, balancing video type, knob count, duration, location and cast.

Claude Opus 5.5 rewrites each description into a video instruction following one specification. The instruction keeps the location, people and camera, removes dialogue and inner states, and adds positions, routes and final states. Single-shot cases last 5 to 10\,s. Multi-shot cases last 15\,s with cuts at 5\,s and 10\,s. Each $k \in \{0,1,2,4,6\}$ has 40 cases, 10 per video type, so a case has 2.6 knobs on average. \cref{fig:prompt_example} shows a third-person case with one knob.

GPT Image 2.0~\cite{openai2026gptimage2} generates the first frame $I_0$ from the opening layout and entity list, plus one frame per later shot in multi-shot cases. The prompt fixes the scene, the camera, the position and pose of every person, the key props and the number of people. A frame that fails a visual check is regenerated once.

\begin{figure}[!htb]
\begin{promptboxwide}[stageCon]{Example Video Instruction}
Opening (0\,s): A rundown modern factory yard on an overcast summer day, with dirt ground and wood-and-sheet-metal sheds (landmark L2). {[}\dots{]} Static eye-level wide shot with horizontal framing, keeping full bodies and both stopping points in frame. P1, a short-haired male reporter in a grey short-sleeved shirt and dark trousers, crouches before the right-hand shutter, {[}\dots{]} holding a black camera O1 at his chest in both hands. P2, a dark-skinned middle-aged worker {[}\dots{]} stands before the left passage facing right, arms down. Exactly 2 people are in frame.

0-2\,s: P1 raises his head from the shutter gap and turns it left. 1-2\,s: he straightens his legs to stand before the shutter and turns his body left, camera still in both hands at his chest.

2-6\,s: P1 walks left {[}\dots{]} and stops before the passage facing left. 2-6\,s: P2 walks right from the left passage across the clear strip, crossing P1 at centre at 4\,s on the side nearer the camera. {[}\dots{]}

6-8\,s (end): P1 stays before the left passage, body and head facing left, camera in both hands at his chest. {[}\dots{]} P2 stays before the right-hand shutter facing right, arms down.
\end{promptboxwide}
\caption{\textbf{Example Video Instruction.}}
\label{fig:prompt_example}
\end{figure}

\section{Evaluation Protocol}
\label{sec:supp_check}\label{sec:supp_levels}\label{sec:supp_l3}\label{sec:supp_judge_prompt}
\cref{tab:supp_l1} lists the seven L1 checks shared by all cases. \cref{tab:supp_l2} lists the four L2 checks of each video type. Each check has written anchors for the scores 0, 0.5 and 1.

Each case has three or four base-scene checks (opening layout, main event, final state) and one check per knob, giving 605 base-scene and 520 knob checks. Each check names the people or props, a time window in whole seconds and the expected state. The checks were drafted from the instruction before any clay or video existed, reviewed by hand and frozen.

Each video is scored in a separate session from an anonymous evidence packet with the checklist, instruction, entity list, first frame and frame sheets, with no metadata naming the source film, agent or clay type. The sheets show every whole second, each check window and each cut. For final videos, each sheet places the clay frame above the final frame, so that G2 can be scored. \cref{fig:prompt_judge} shows an excerpt of the judge prompt.

\begin{table*}[t]
\centering
\caption{\textbf{L1 Universal checks.}}
\label{tab:supp_l1}
\footnotesize\setlength{\tabcolsep}{4pt}
\begin{tabular}{@{}l l p{0.72\linewidth}@{}}
\toprule
ID & Check & Criterion \\
\midrule
G1 & Instruction adherence & The scene, character roles, main actions and phases follow the instruction. \\
G2 & Following the clay & The final video follows the visible clay positions, paths, framing and event order. Clay errors are logged separately and the text does not override the clay. \\
G3 & Closed world & The listed main people and props are kept without additions, duplicates, mergers or unjustified deletions. Normal occlusion and reflections do not add people. \\
G4 & Identity stability & People and props keep a recognizable identity and shape across frames without unjustified flicker, transformation or drift. \\
G5 & Physical plausibility & Feet, hands, object contact and occlusion are physically plausible. Only exceptional behaviour stated in the instruction is permitted. \\
G6 & Visual clarity & The final video is clear, without broken contours, ghosting or flicker that obscures people, motion or required readable text. \\
G7 & Finished appearance & The final video has the specified finished appearance instead of grey human or geometric proxy materials. \\
\bottomrule
\end{tabular}
\end{table*}

\begin{table*}[t]
\centering
\caption{\textbf{L2 Categorical checks per video type.}}
\label{tab:supp_l2}
\footnotesize\setlength{\tabcolsep}{4pt}
\begin{tabular}{@{}l l p{0.72\linewidth}@{}}
\toprule
Type & Check & Criterion \\
\midrule
\multirow{4}{*}{Third-person} & Camera motion & The camera keeps the fixed framing or the specified viewing direction and shot-scale changes. The final video follows the visible clay camera. \\
 & Depth ordering & People, furniture and props keep visible front and back occlusion relations without passing through solid objects. \\
 & Body support & People keep proportions suited to the environment, with grounded feet and plausible contact and support. \\
 & Background stability & Walls, floor, furniture and fixed props keep their layout through fixed framing or camera motion. \\
\midrule
\multirow{4}{*}{First-person} & Viewpoint & The view stays at the operator's eye position, with only the permitted arms or body parts in frame and no switch to a third-person view. \\
 & Hand-tool continuity & The number of hands, arms and tools and their grip states stay continuous, without extra limbs or tools. \\
 & Head motion & The viewpoint stays fixed or completes the specified head turns, movement and looking back. The final video follows the clay viewpoint. \\
 & Manipulated objects & Hands contact, pick up or place the correct props, and objects do not jump into the hands from wrong positions. \\
\midrule
\multirow{4}{*}{Multi-shot} & Cut timing & The video has three shots with clear cuts at 5\,s and 10\,s, each showing the prescribed viewpoint. \\
 & Cross-shot identity & People keep identity and role across cuts. The final video also keeps clothing and appearance. \\
 & Spatial layout & Landmark, person and prop relations after each cut match the specified viewpoints. Normal reverse-shot changes do not count as layout changes. \\
 & Action continuity & Actions and prop states continue across cuts without unjustified resets. \\
\midrule
\multirow{4}{*}{2D animation} & Flat style & The video keeps flat side-view animation, with people and props moving within one 2D frame. \\
 & Text and symbols & Required readable text, numbers, symbols and interface states are clear, correct and stable. \\
 & Shape consistency & People, flat graphics, props and fixtures keep their prescribed 2D layout relations. \\
 & Order of state changes & Prescribed planar actions, counts and state changes occur in order, and completed states do not reset. \\
\bottomrule
\end{tabular}
\end{table*}

\begin{figure}[!htb]
\begin{promptboxwide}[stageEval]{Judge Prompt}
You are an independent, anonymous visual judge. Rate only the anonymous candidate in this packet. First read the protocol, the manifest, the checklist, the task text and the entity list, then personally view every image listed in the manifest. {[}\dots{]} Each rationale must state what is seen and at which absolute second.

\plabel{Scores} Rate each check against its frozen anchors:

1: fully satisfied. 0.5: clearly partially satisfied. 0: visible and not satisfied. na: visible evidence is insufficient.

For each item write one sentence stating the absolute second and the observed fact. A requirement may not be reported as an observed fact.

For every shot, view its first frame and the frames just before, at and after each hard cut. The correctness of a cut may not be inferred from middle frames.
\end{promptboxwide}
\caption{\textbf{Judge Prompt.}}
\label{fig:prompt_judge}
\end{figure}

\subsection{Human evaluation}
\label{sec:supp_human}
We randomly draw 500 cases, each with one clay video and one final video, balanced over the video generation models and clay types, so each annotator rates 1{,}000 videos. Three annotators, volunteers from our lab who gave informed consent, see the video, the instruction and the first frame and rate L1, L2 and L3 from 1 (not realized) to 5 (fully realized). Videos carry anonymous IDs in a random order per annotator, with the clay and final video of a case never adjacent. Annotators see no checklist or judge score and rate independently. Krippendorff's $\alpha$~\cite{krippendorff2004reliability} is 0.79 on clay videos and 0.76 on final videos (\cref{tab:human}). No personal data was collected.

\section{Pipeline Details}
\label{sec:supp_pipe}\label{sec:supp_clay}\label{sec:supp_agents}\label{sec:supp_vgm}
\begin{figure*}[t]
\centering
\includegraphics[width=\linewidth]{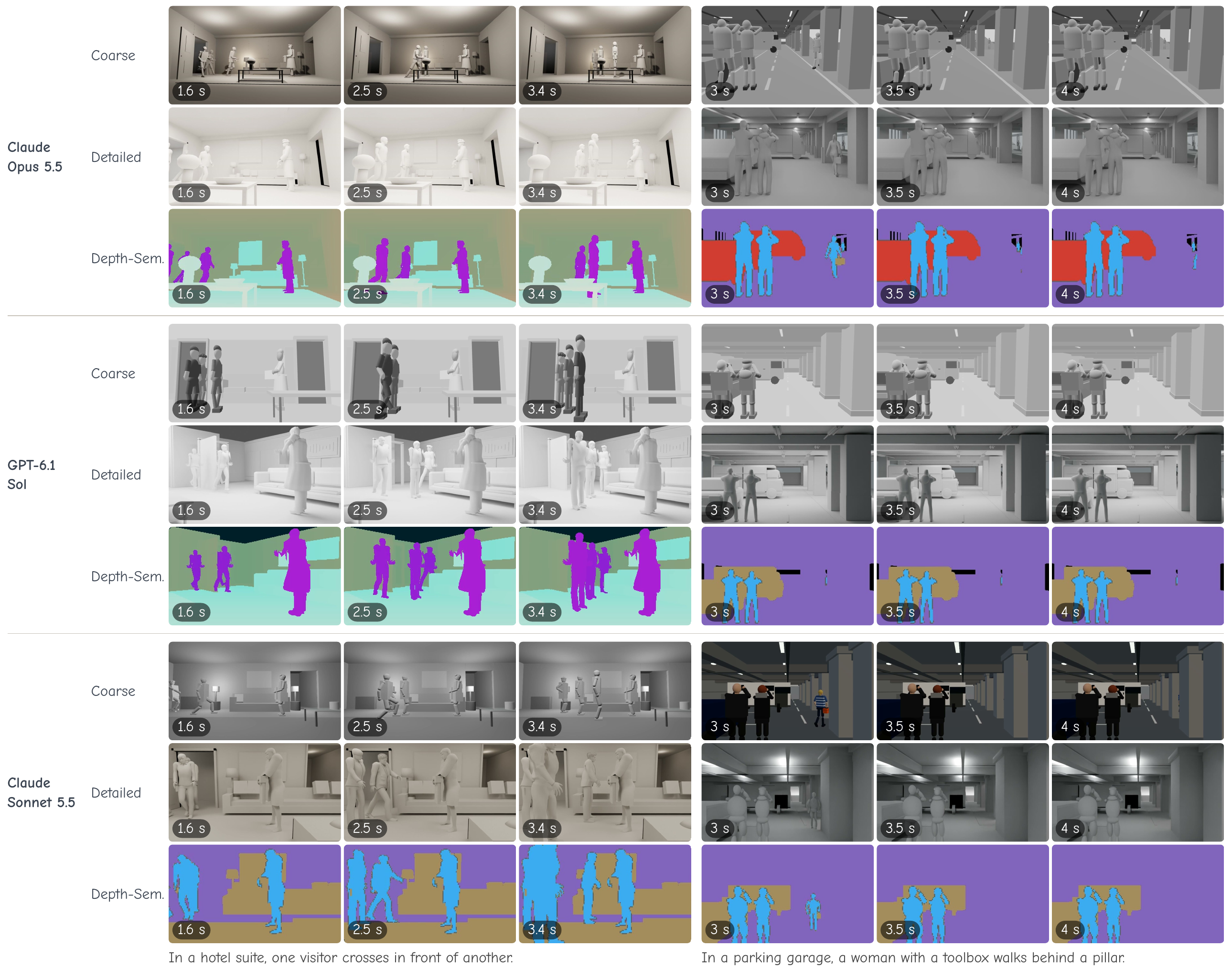}
\caption{\textbf{Coarse, Detailed and Depth-Semantic clay} for two cases built by three coding agents. Depth-Semantic panels show the semantic map. Agents may color their clay, as Claude Sonnet 5.5 does in the garage case.}
\label{fig:supp_claytypes}
\end{figure*}

\cref{fig:supp_claytypes} shows the three clay representations built by each coding agent. \textbf{Coarse} clay is a Three.js~\cite{cabello2010threejs} program with primitive bodies on an absolute timeline. \textbf{Detailed} clay is a Blender~\cite{blender2024} Python program with rigged MPFB2 bodies retargeted to CMU motion capture, rendered at $768\times432$ and 24\,fps. \textbf{Depth-Semantic} clay is exported from the Detailed scene as per-frame depth and semantic maps in the Code World Model~\cite{chen2026codeworld} format, at no extra coding cost.

GPT-6.1 Sol writes one scene plan per case with metric coordinates, landmarks, start poses, props, one camera per shot and a timed list of events. Every coding agent receives the same plan, instruction $T$ and first frames $I_0$. Claude Opus 5.5 and Claude Sonnet 5.5 run in Claude Code and GPT-6.1 Sol in Codex. A renderer checks duration, frame rate, resolution and empty frames, and a technical failure allows one retry. Content errors are not repaired.

All video generation models receive the same prompt, with the clay video as the structure reference and the first frame as the appearance reference. Multi-shot cases are generated in one 15\,s pass. The \texttt{I2V} prompt drops only the sentences on the clay (\cref{fig:prompt_vgm}).

\begin{figure*}[t]
\begin{promptboxwide}[stageCtoV]{Video Generation Prompt}
\plabel{With clay, \texttt{C2V}}

subject\_definitions: \textless{}Video 1\textgreater{} is the grey clay reference fixing layout, paths, timing and camera. \textless{}Picture 1\textgreater{} is the opening appearance reference. \textless{}Subject 1\textgreater{} P1: short-haired adult male reporter, grey short-sleeved shirt and dark trousers, holding a small black camera in both hands. \textless{}Subject 2\textgreater{} P2: dark-skinned middle-aged male worker, short black hair, blue-and-white striped short-sleeved polo shirt and black trousers.

summary: A live-action reporter rises beside a rusted factory shutter and walks left across a dirt yard as a striped-shirt worker crosses right. Duration: 8 seconds {[}\dots{]} follow \textless{}Video 1\textgreater{} for geometry and timing and \textless{}Picture 1\textgreater{} for appearance.

retention\_analysis: \textless{}Video 1\textgreater{}: fully\_preserved for layout, paths, camera and timing {[}\dots{]}. \textless{}Picture 1\textgreater{}: partially\_preserved for character appearance and scene styling {[}\dots{]}. Preserve identities and the closed cast.

detailed\_description: Opening (0\,s): A rundown modern factory yard on an overcast summer day, with dirt ground and wood-and-sheet-metal sheds L2. {[}\dots{]} Exactly 2 people are in frame. 0-2\,s: {[}\dots{]} 2-6\,s: \textless{}Subject 2\textgreater{} walks right from the left passage across the clear strip, crossing \textless{}Subject 1\textgreater{} at centre at 4\,s on the side nearer the camera. Both heads and shoulders remain separately visible without bodily contact. {[}\dots{]} 6-8\,s (end): \textless{}Subject 1\textgreater{} stays before the left passage, body and head facing left, camera in both hands at his chest {[}\dots{]}.
\tcblower
\plabel{Without clay, \texttt{I2V}}

subject\_definitions: \textless{}Picture 1\textgreater{} is the opening appearance reference. \textless{}Subject 1\textgreater{} P1: short-haired adult male reporter, grey short-sleeved shirt and dark trousers, holding a small black camera in both hands. \textless{}Subject 2\textgreater{} P2: dark-skinned middle-aged male worker {[}\dots{]}.

summary: A live-action reporter rises beside a rusted factory shutter and walks left across a dirt yard as a striped-shirt worker crosses right. Duration: 8 seconds {[}\dots{]} follow \textless{}Picture 1\textgreater{} for appearance.

retention\_analysis: \textless{}Picture 1\textgreater{}: partially\_preserved for character appearance and scene styling {[}\dots{]}. Preserve identities and the closed cast.

detailed\_description: identical to the \texttt{C2V} prompt.
\end{promptboxwide}
\caption{\textbf{Video Generation Prompt.}}
\label{fig:prompt_vgm}
\end{figure*}

\section{Full Results}
\label{sec:supp_results}
\begin{table*}[t]
\centering
\caption{\textbf{Scores (\%) by video type} for Claude Opus 5.5 clays, two-judge mean. All is the mean of L1 to L3.}
\label{tab:supp_type_full}
\footnotesize
\begin{tabular*}{\textwidth}{@{\extracolsep{\fill}}l cccc cccc cccc cccc@{}}
\toprule
 & \multicolumn{4}{c}{Third-person} & \multicolumn{4}{c}{First-person} & \multicolumn{4}{c}{Multi-shot} & \multicolumn{4}{c}{2D animation} \\
\cmidrule(lr){2-5}\cmidrule(lr){6-9}\cmidrule(lr){10-13}\cmidrule(lr){14-17}
Clay input & L1 & L2 & L3 & All & L1 & L2 & L3 & All & L1 & L2 & L3 & All & L1 & L2 & L3 & All \\
\midrule
\multicolumn{17}{l}{\textit{\lgMM MiniMax-H3}} \\
\hspace{0.6em}I2V & \textbf{83.5} & 89.7 & 64.4 & 79.2 & \textbf{83.4} & 82.4 & 62.8 & 76.2 & \textbf{81.0} & \textbf{80.2} & 72.7 & \textbf{78.0} & \textbf{84.2} & 77.3 & 65.2 & 75.6 \\
\hspace{0.6em}Coarse & 82.2 & 90.0 & \textbf{72.1} & \textbf{81.4} & 80.9 & 86.7 & 67.3 & 78.3 & 74.5 & 78.5 & 71.8 & 74.9 & 80.7 & 80.0 & 72.2 & 77.6 \\
\hspace{0.6em}Detailed & 81.7 & 90.7 & 71.1 & 81.2 & 80.8 & 87.4 & \textbf{67.8} & 78.7 & 78.6 & 78.3 & 72.2 & 76.4 & 80.6 & 79.2 & 72.5 & 77.4 \\
\hspace{0.6em}Depth-Sem. & 82.2 & \textbf{90.8} & 70.8 & 81.3 & 82.2 & \textbf{87.5} & 67.5 & \textbf{79.1} & 78.2 & 79.2 & \textbf{72.8} & 76.7 & 83.3 & \textbf{80.4} & \textbf{72.8} & \textbf{78.8} \\
\midrule
\multicolumn{17}{l}{\textit{\lgSD Seedance 2.0}} \\
\hspace{0.6em}I2V & 84.6 & 91.0 & 68.2 & 81.3 & 83.8 & 88.3 & 66.4 & 79.5 & \textbf{82.1} & 79.8 & 67.1 & 76.3 & 83.8 & 77.0 & 64.8 & 75.2 \\
\hspace{0.6em}Coarse & 85.2 & 93.6 & 81.9 & 86.9 & 84.2 & 93.0 & 78.1 & 85.1 & 76.8 & 81.8 & 77.8 & 78.8 & 85.9 & 86.3 & 81.7 & 84.6 \\
\hspace{0.6em}Detailed & 86.3 & 93.8 & \textbf{83.5} & \textbf{87.9} & 84.7 & 92.1 & \textbf{79.2} & 85.3 & 79.1 & 83.6 & \textbf{80.2} & \textbf{81.0} & 86.3 & 85.5 & 80.3 & 84.0 \\
\hspace{0.6em}Depth-Sem. & \textbf{87.4} & \textbf{95.7} & 79.8 & 87.6 & \textbf{85.7} & \textbf{94.6} & 77.2 & \textbf{85.8} & 81.0 & \textbf{85.1} & 75.6 & 80.6 & \textbf{88.2} & \textbf{87.9} & \textbf{82.7} & \textbf{86.3} \\
\midrule
\multicolumn{17}{l}{\textit{\lgKL Kling 4.0}} \\
\hspace{0.6em}I2V & \textbf{85.8} & 91.7 & 68.3 & 81.9 & \textbf{84.4} & 88.2 & 67.0 & 79.9 & \textbf{82.5} & 79.0 & 71.0 & 77.5 & \textbf{87.4} & 80.2 & 70.7 & 79.4 \\
\hspace{0.6em}Coarse & 82.7 & 90.5 & 79.2 & 84.1 & 81.1 & 90.7 & 76.1 & 82.6 & 77.6 & 79.7 & 73.0 & 76.8 & 86.8 & \textbf{85.7} & \textbf{80.8} & \textbf{84.4} \\
\hspace{0.6em}Detailed & 83.9 & 91.5 & \textbf{81.0} & 85.5 & 82.3 & 91.1 & \textbf{78.8} & \textbf{84.1} & 77.7 & 79.3 & \textbf{74.6} & 77.2 & 84.5 & 84.3 & 77.3 & 82.0 \\
\hspace{0.6em}Depth-Sem. & 84.7 & \textbf{92.8} & 79.2 & \textbf{85.6} & 83.2 & \textbf{92.2} & 75.0 & 83.5 & 78.9 & \textbf{83.5} & 74.5 & \textbf{79.0} & 86.3 & 83.2 & 80.1 & 83.2 \\
\bottomrule
\end{tabular*}
\end{table*}

\cref{tab:supp_type_full} gives the scores by video type. Rows give, for each video generation model, the final videos from \texttt{I2V} and from the three clay representations. Columns give L1, L2, L3 and All for each of the four video types. All clays are built by Claude Opus 5.5, and each score is the two-judge mean.

\section{Qualitative Examples}
\label{sec:supp_qual}
\begin{figure*}[!tp]
\centering
\includegraphics[width=0.75\linewidth]{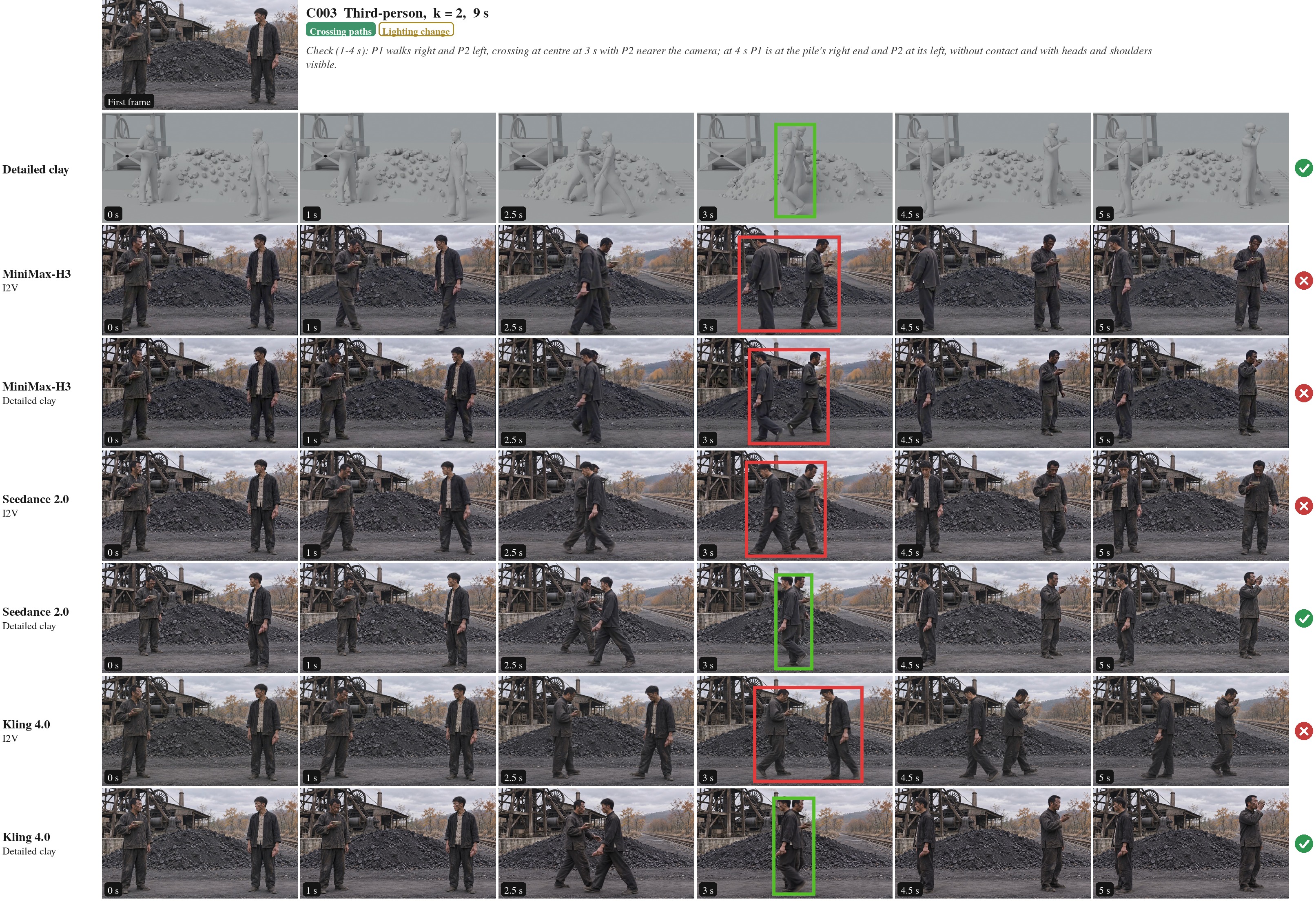}
\caption{\textbf{Crossing paths in a third-person video ($k{=}2$).} C003: the clay places the crossing of P1 and P2 at the centre at 3\,s. Seedance 2.0 and Kling 4.0 with Detailed clay follow it. Without clay, Seedance 2.0 crosses before 3\,s and Kling 4.0 has not crossed at 3\,s. MiniMax-H3 crosses before 3\,s with and without clay.}
\label{fig:gal_c003}
\end{figure*}

\begin{figure*}[!tp]
\centering
\includegraphics[width=0.75\linewidth]{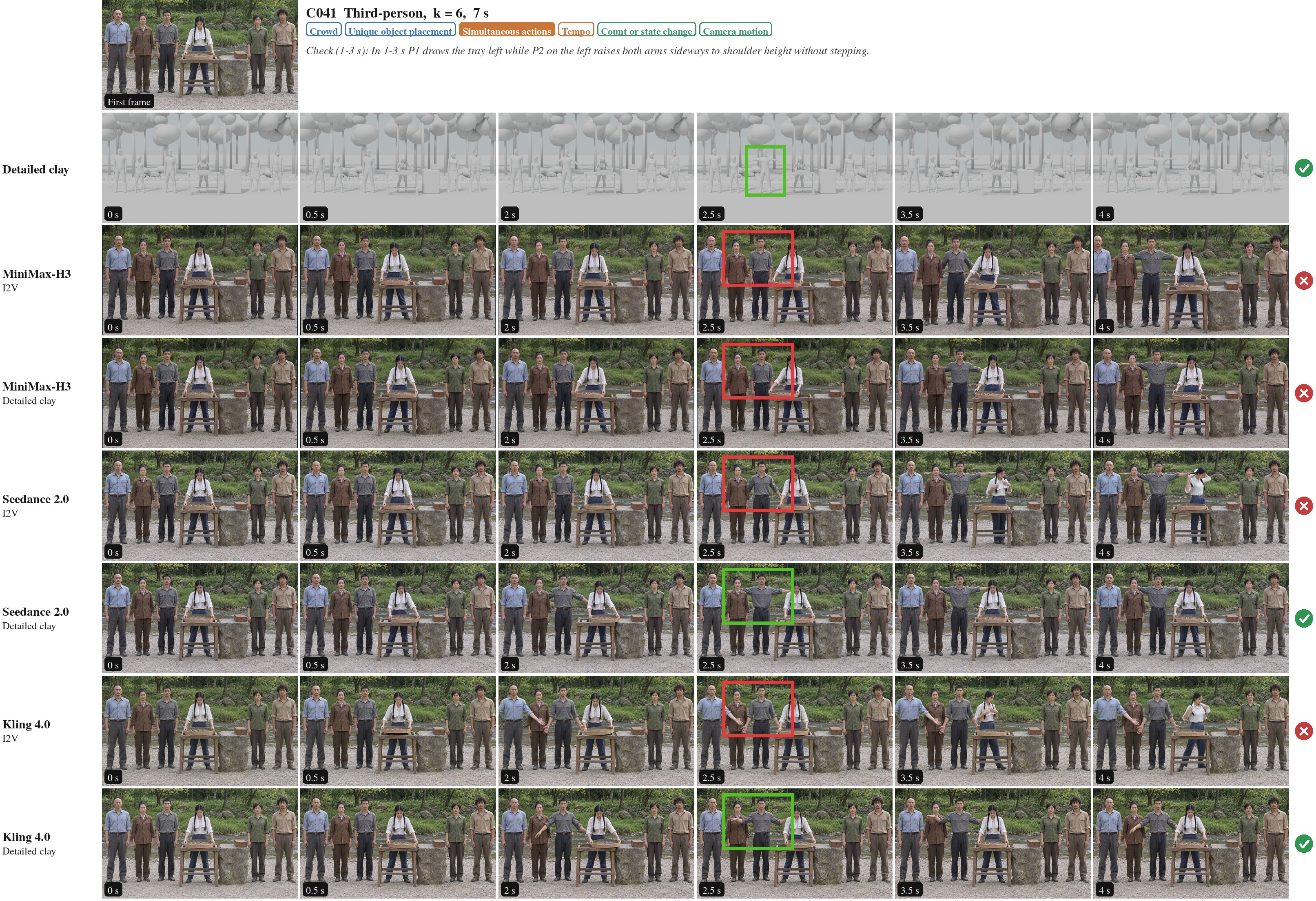}
\caption{\textbf{Simultaneous actions in a third-person video ($k{=}6$).} C041: while P1 draws the tray left in 1-3\,s, P2 raises both arms sideways. Seedance 2.0 and Kling 4.0 with Detailed clay raise the arms by 2.5\,s, as in the clay. All three I2V videos and MiniMax-H3 with Detailed clay keep the arms down at 2.5\,s.}
\label{fig:gal_c041}
\end{figure*}

\begin{figure*}[!tp]
\centering
\includegraphics[width=0.858\linewidth]{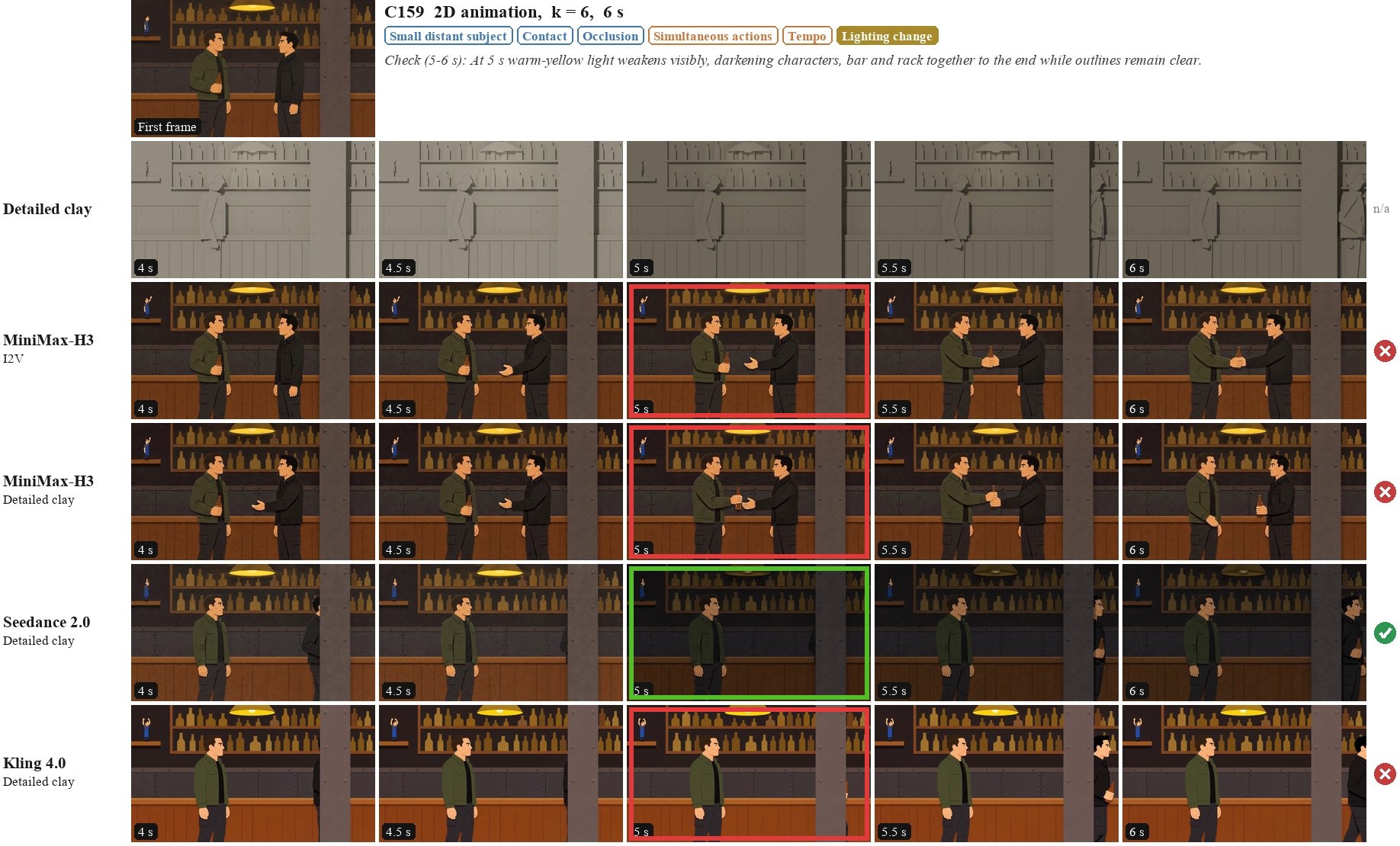}
\caption{\textbf{Dimming the light in a 2D animation ($k{=}6$).} C159, Lighting change: Seedance 2.0 with Detailed clay dims the warm bar light at 5 s and keeps the people, bar and rack dark to the end. MiniMax-H3 I2V, MiniMax-H3 with Detailed clay and Kling 4.0 with Detailed clay keep the original brightness through 6 s. Lighting is not scored on the clay.}
\label{fig:gal_hero_02}
\end{figure*}

\begin{figure*}[!tp]
\centering
\includegraphics[width=0.706\linewidth]{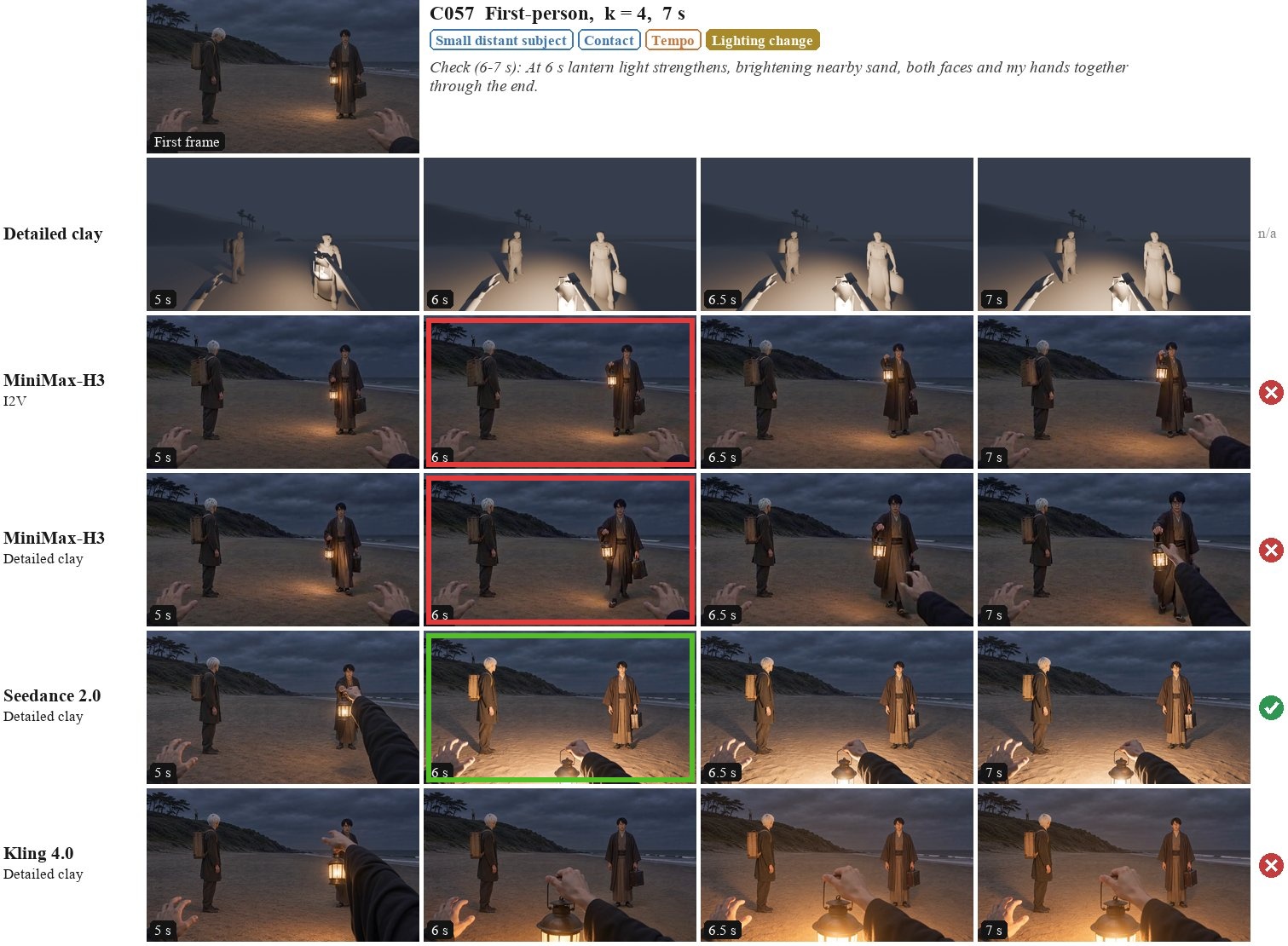}
\caption{\textbf{Brightening the scene in a first-person video ($k{=}4$).} C057, Lighting change: Seedance 2.0 with Detailed clay brightens the sand and faces at 6 s as the lantern reaches the viewer's hands. In MiniMax-H3 with Detailed clay the lantern stays in P2's hand and the scene stays dark, and MiniMax-H3 I2V shows no clear change at 6 s. Kling 4.0 with Detailed clay brightens the scene, but one judge gives only half credit.}
\label{fig:gal_hero_03}
\end{figure*}

\begin{figure*}[!tp]
\centering
\includegraphics[width=1.000\linewidth]{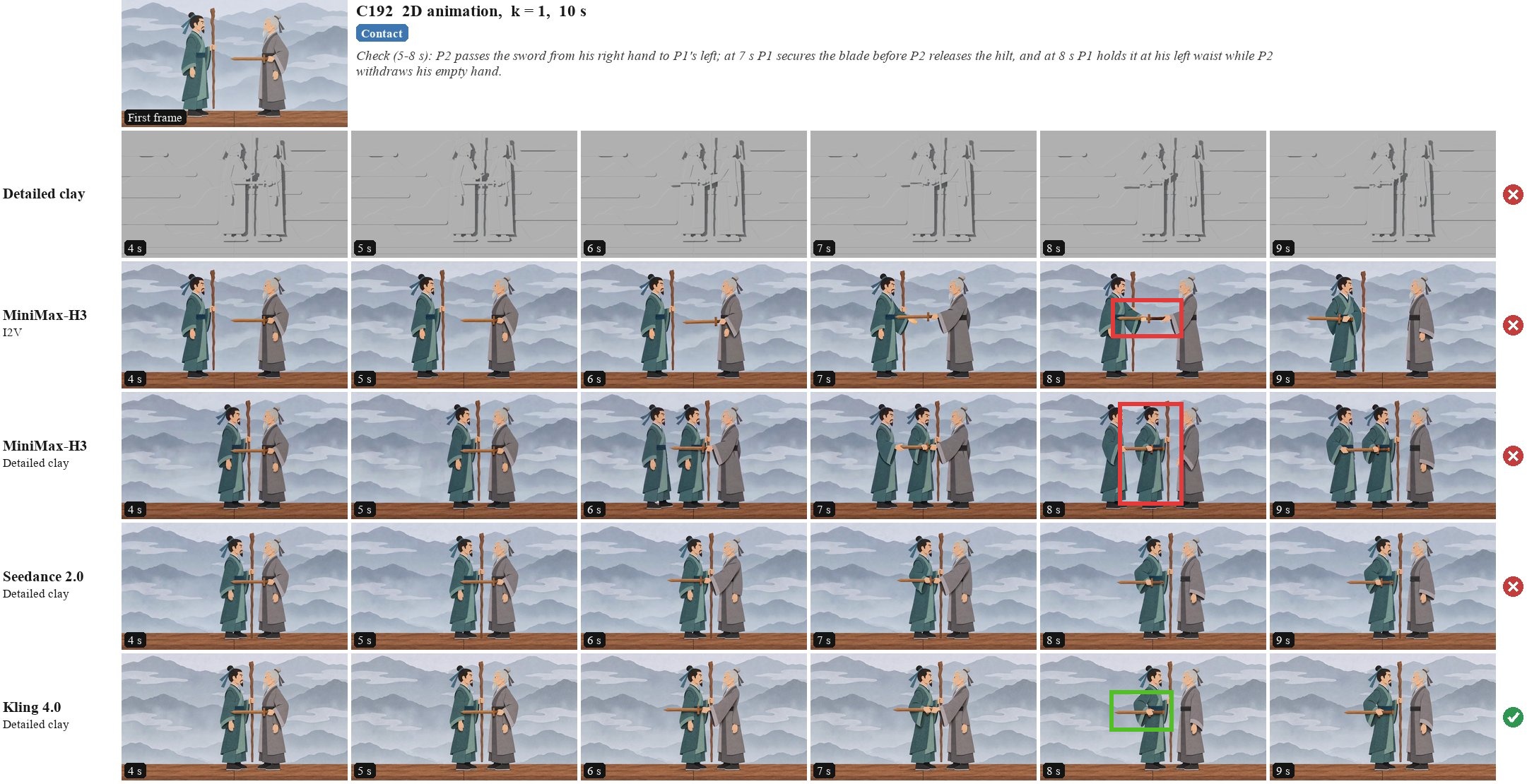}
\caption{\textbf{Handing over a sword in a 2D animation ($k{=}1$).} C192, Contact: Kling 4.0 with Detailed clay passes the sword from P2 to P1 at 7 s, and P1 holds it at his waist by 8 s. MiniMax-H3 with Detailed clay adds a third figure who takes the sword. MiniMax-H3 I2V, Seedance 2.0 with Detailed clay and the clay show the handover only partly.}
\label{fig:gal_hero_04}
\end{figure*}

\begin{figure*}[!tp]
\centering
\includegraphics[width=1.000\linewidth]{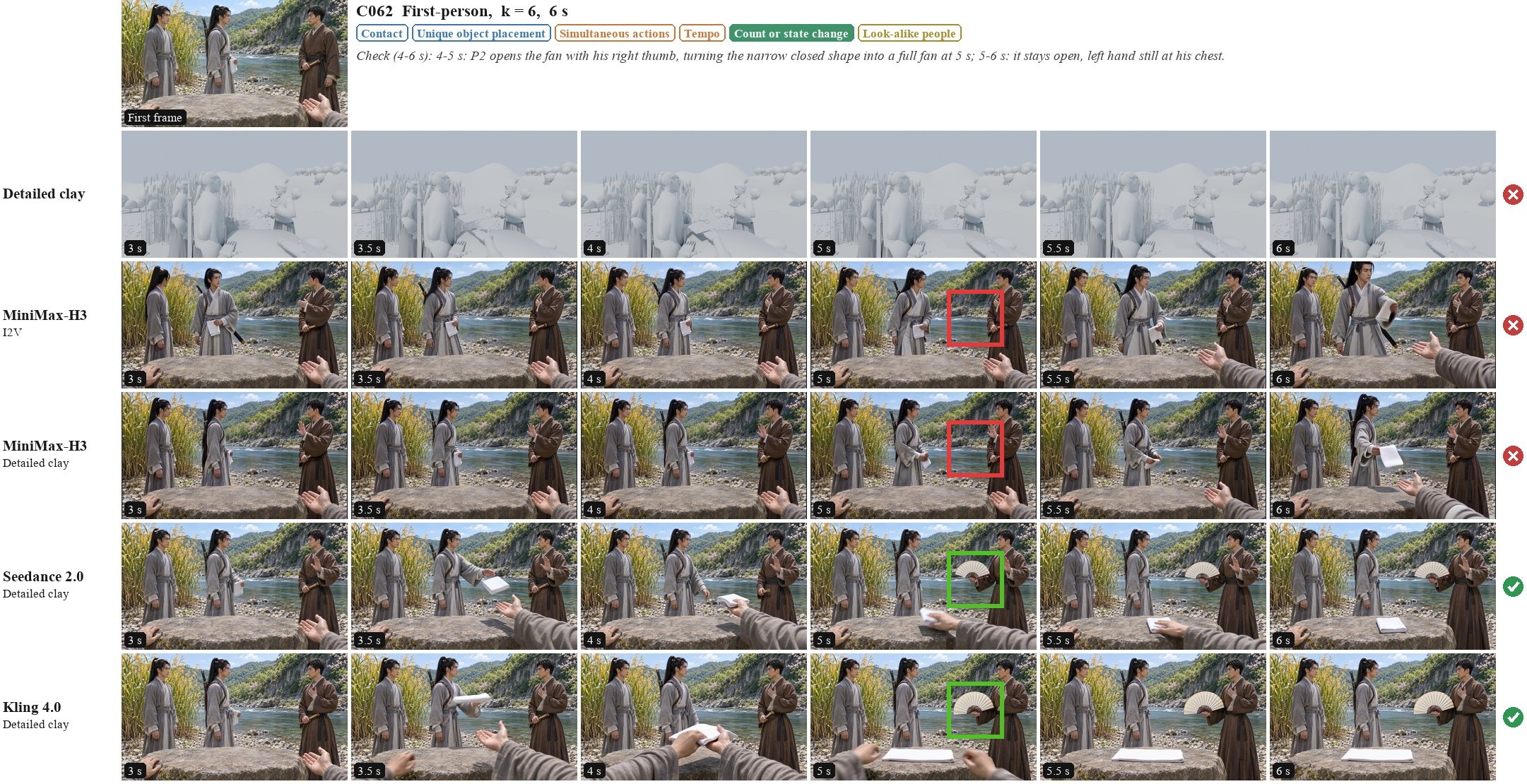}
\caption{\textbf{Opening a fan in a first-person video ($k{=}6$).} C062, Count or state change: Seedance 2.0 and Kling 4.0 with Detailed clay open the fan into a full fan by 5 s. MiniMax-H3 I2V and MiniMax-H3 with Detailed clay never perform this action, and the fan stays closed in P2's hand through 6 s.}
\label{fig:gal_hero_05}
\end{figure*}

\begin{figure*}[!tp]
\centering
\includegraphics[width=0.706\linewidth]{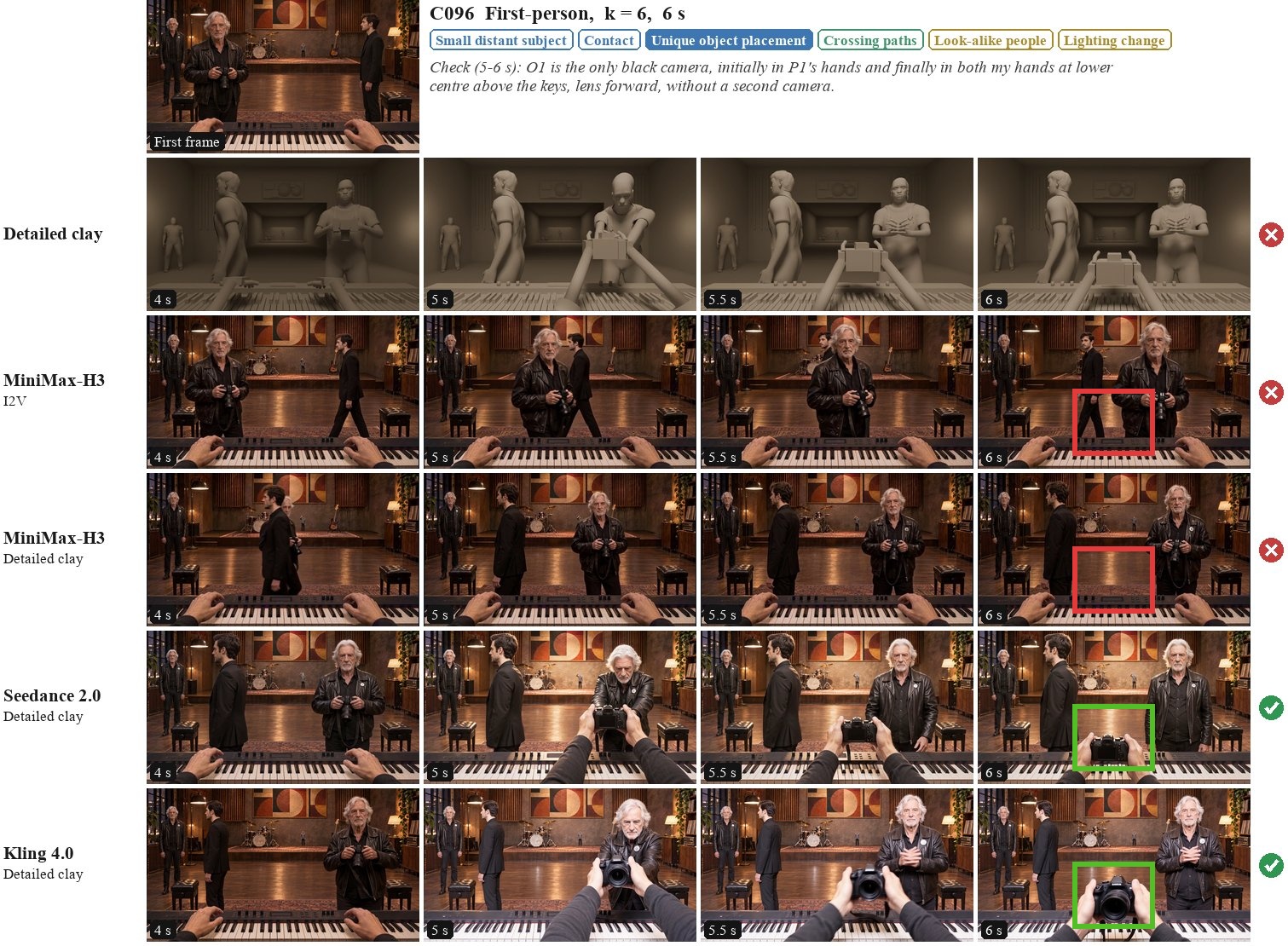}
\caption{\textbf{Handing a camera to the viewer in a first-person video ($k{=}6$).} C096, Unique object placement: Seedance 2.0 and Kling 4.0 with Detailed clay hand the camera to the viewer's hands at 5 s and hold it above the keys to 6 s. MiniMax-H3 I2V and MiniMax-H3 with Detailed clay leave the camera in the wrong place, at P1's chest through 6 s.}
\label{fig:gal_hero_06}
\end{figure*}

Each of \cref{fig:gal_c003} to \cref{fig:gal_hero_06} shows one case: the first frame, the active knobs with the checked one filled, one L3 check, and frames across its window for the clay and the final videos. Row labels give the video generation model and its input. A green tick means both judges give 1, a red cross marks any other score and a grey n/a marks a check not scored on clay. Green and red boxes mark where the check passes or fails. \Cref{fig:gal_c003,fig:gal_c041} also show the \texttt{I2V} video of each video generation model.

\end{document}